\documentclass[twoside]{article}
\usepackage[accepted]{aistats2018}
\usepackage[utf8]{inputenc} 
\usepackage[T1]{fontenc}    
\usepackage{lmodern}

\usepackage{balance}

\usepackage{booktabs}       
\usepackage{amsfonts}       
\usepackage{nicefrac}       
\usepackage{microtype}      

\usepackage{amsmath,amsfonts,amssymb,amsthm}
\usepackage{graphicx}
\usepackage{wrapfig}
\usepackage{wasysym}
\usepackage{enumitem}
\usepackage{epstopdf}
\usepackage[ruled,linesnumbered]{algorithm2e}

\usepackage[percent]{overpic}

\usepackage{color}
\usepackage{xcolor}
\definecolor{orange}{rgb}{0,0,0.6} 

\newcommand{\myparagraph}[1]{\noindent\textbf{#1.}}
\newtheorem{thm}{Theorem}

\newtheorem{prop}{Proposition}
\newtheorem{remark}{Remark}

\def\X{\mathbf{X}}

\def\U{\mathbf{U}}
\def\V{\mathbf{V}}
\def\u{\mathbf{u}}
\def\v{\mathbf{v}}
\def\r{\mathbf{r}}
\def\A{\mathbf{A}}

\def\S{\mathcal{S}}
\def\1{\mathbf{1}}
\def\0{\mathbf{0}}
\newcommand{\opt}{{\rm opt}}
\newcommand{\Omd}{\Omega_{\scalebox{.6}{dropout}} }
\newcommand{\Bt}{ {\rm Bernoulli}(\theta) }
\newcommand{\eg}{\emph{e.g.}}

\newcommand{\diag}[1]{{\rm diag}(#1)}

\begin{document}

%

%

\twocolumn[

\aistatstitle{Dropout as a Low-Rank Regularizer for Matrix Factorization}

\aistatsauthor{ Jacopo Cavazza$^1$, Pietro Morerio$^1$, Benjamin Haeffele$^2$, Connor Lane$^2$, Vittorio Murino$^1$, Ren\'{e} Vidal$^2$}

\aistatsaddress{ $^1$ Pattern Analysis and Computer Vision, Istituto Italiano di Tecnologia, Genova, 16163, Italy \\ $^2$ Johns Hopkins University, Center for Imaging Science, Baltimore, MD 21218, USA \\ \\ \texttt{jacopo.cavazza@iit.it}, \texttt{connor.lane@jhu.edu}, \texttt{bhaeffele@jhu.edu}, \\ \texttt{vittorio.murino@iit.it}, \texttt{rvidal@cis.jhu.edu}  } ]

\begin{abstract}
	Regularization for matrix factorization (MF) and approximation problems has 
	been carried out in many different ways. Due to its popularity in deep learning, 
	dropout has been applied also for this class of problems. Despite its solid empirical
	performance, the theoretical properties of dropout as a regularizer remain quite 
	elusive for this class of problems. 
	In this paper, we present a theoretical analysis of dropout for MF, where Bernoulli random 
	variables are used to drop columns of the factors. We demonstrate the 
	equivalence between dropout and a fully deterministic model for MF in which the 
	factors are regularized by the sum of the product of squared Euclidean norms of 
	the columns. Additionally, we inspect the case of a variable sized factorization 
	and we prove that dropout achieves the global minimum of a convex 
	approximation problem with (squared) nuclear norm regularization. As a result, we conclude that dropout can be used as a low-rank regularizer with data dependent singular-value thresholding.
\end{abstract}

\section{INTRODUCTION}\label{sez:intro}

In many problems in machine learning and artificial intelligence, no matter what 
the input dimensionality of the raw data is, relevant patterns and information 
often lie in a low-dimensional manifold. In order to capture its structure, 
linear subspaces have become very popular, arguably due to their efficiency and 
versatility \cite{Linspace_survey}. 

Mathematically, a linear subspace is obtainable from data points 
$\mathbf{x}_1,\dots,\mathbf{x}_m \in \mathbb{R}^n$ as follows. 
We build the $m \times n$ matrix $\X$ that stacks each sample by rows. Then, 
when 
looking for a $d$-dimensional embedding, we search for two matrices, $\U \in 
\mathbb{R}^{m \times d}$ and $\V \in \mathbb{R}^{n \times d}$, such that 
$\X 
\approx \U \V^\top$. Algorithmically, $\U$ and $\V$ can be found through 
optimization, according to the \textit{matrix factorization} (MF) problem
\begin{equation}\label{eq:MF}
\min_{\U,\V} \| \X - \U \V^\top  \|_F^2 + \lambda \Omega(\U,\V)
\end{equation}
where the Frobenius norm is a well established proxy to impose similarity 
between $\X$ and $\U \V^\top$. Also, for $\lambda > 0$, the regularizer $  
\Omega(\U,\V)$ in \eqref{eq:MF} imposes some constraints on 
the factors: for instance, orthonormality as in PCA \cite{Vidal:book}.

Two are the main advantages of \eqref{eq:MF}. First, we optimize on the factors 
directly, achieving a structured decomposition of $\X$. Second, the number of 
variables to be optimized scales linearly with respect to $m + n$, ensuring 
applicability even in the big data regime. Unfortunately, a big shortcoming 
in \eqref{eq:MF} arises. Indeed, when $\U$ is fixed, optimizing for $\V$ is a convex 
problem and vice versa, but, \eqref{eq:MF} is \underline{not} convex when 
optimizing on $\U$ and 
$\V$ jointly. Therefore, one needs ancillary optimality conditions to ensure 
that the global optimum $(\U^\opt,\V^\opt)$ of \eqref{eq:MF} exists as well as algorithms to compute a global optimum \cite{H14,H15,H17}.

Those issues can be solved by replacing the MF problem \eqref{eq:MF} 
with \emph{matrix approximation}, that is,
\begin{equation}\label{eq:MA}
\min_{\mathbf{A}} \| \X - \A \|_F^2 + \gamma \Xi(\A).
\end{equation}
In \eqref{eq:MA}, $\gamma > 0$ and we minimize over $\A \in \mathbb{R}^{m 
\times n}$, forcing it to be close enough to $\X$ after adding the penalization 
term $\Xi$ which plays the analogous role on $\A$ as $\Omega$ does on 
$\U$ and $\V$ in \eqref{eq:MF}. 

The formulations in \eqref{eq:MF} and \eqref{eq:MA} are highly complementary. For instance, differently from \eqref{eq:MF}, the optimization in 
\eqref{eq:MA} is convex and therefore, a global minimizer exists, is unique and 
can be found via gradient descent (and, sometimes, it has a closed-form solution, \eg, when $\Xi 
= \| \cdot \|^2_F$). Again, differently from \eqref{eq:MF}, the problem in 
\eqref{eq:MA} is not scalable (due to the $m\cdot n$ variables to be optimized) 
and, also, the optimal solution $\A^\opt$ of \eqref{eq:MA} does not have the structure 
that \eqref{eq:MF} provides in terms of explicit factors $\U$ and $\V$.

In this paper, we bridge the gap between factorization 
\eqref{eq:MF} and approximation \eqref{eq:MA} for matrices, ultimately 
providing an unified framework by means of a recently developed strategy from 
deep learning: \emph{dropout}.

Dropout \cite{DropoutCORR,DropoutJMLR} is a popular algorithm for training neural 
networks while preventing overfitting. During dropout training, each unit is 
endowed with a (binary) Bernoulli random variable of expected value $\theta$ - 
which is called ``retain probability''. So, for each example/mini-batch, the 
network's weights are updated by using a back-propagation step which only 
involves 
the units whose corresponding Bernoulli variables are sampled with value 1. 
At each iteration, those Bernoulli variables are re-sampled again and the weights are updated accordingly. Note that, since all the 
sub-networks are sampled from the original architecture, the weights are shared 
across different units' subsamplings and dropout can be interpreted 
as a model ensemble. During inference, no units' suppression is performed and, 
simply, all the weights are rescaled by $\theta$, the latter stage being interpreted 
as a model average up to certain approximations \cite{DropoutJMLR,Baldi1,Baldi2}.

Motivated by the significant efforts made to understand dropout as (implicit) regularization \cite{Wager:NIPS13,Baldi1,Baldi2, Gal2016Dropout}, as in \cite{Zhai:CoRR15,He2016}, we combine dropout and MF through the following problem. While still looking for a direct optimization of $\X \approx \U \V^\top$ over factors $\U \in \mathbb{R}^{m \times d}$ and $\V^{n \times d}$, we 
replace \eqref{eq:MF} with 
\begin{equation}\label{eq:dropoutMF}
\min_{\U,\V} \mathbb{E}_\r \left \| \mathbf{X} - \tfrac{1}{\theta} 
\mathbf{U}{\rm 
	diag}(\mathbf{r})\mathbf{V}^\top \right\|_F^2
\end{equation} 
where $\| \cdot \|_F$ denotes the Frobenius norm of a matrix
$\mathbf{r} \in \mathbb{R}^d$ is a random vectors whose entries are i.i.d. $\Bt$, and $E_\mathbf{r}$ denotes the expected value with respect to $\mathbf{r}$. Essentially, by taking directly inspiration from the idea of suppressing ``units'' in a neural network, we here suppress ``columns'' of the factorization in order to obtain an optimization scheme that mimics the actual dropout training for neural networks. Indeed, in neural network training, batches of data are shaped as matrices and, when dropout is applied to the input layer, some columns of that matrix are set to zero. In practice, dropout for MF has shown solid performance \cite{Zhai:CoRR15,He2016}, but, it is still unclear what sort of regularization it induces for such class of problems.


The contributions of the paper are the following:

\begin{enumerate}[leftmargin=*]
\item We demonstrate that dropout for MF \eqref{eq:dropoutMF} is equivalent to the following deterministic regularization framework
\begin{equation}
\min_{\U,\V} \left[ \| \mathbf{X} - \mathbf{U}  \mathbf{V}^\top \|_F^2 + \tfrac{1 - \theta}{\theta} \Omd(\U,\V) \right]
\end{equation}
where
\begin{equation}
\Omd = \sum_{k = 1}^d \| \mathbf{u}_k \|_2^2 \| \mathbf{v}_k \|_2^2.
\end{equation}

\item While carefully inspecting the nature of 
$\Omega_{\scalebox{.6}{dropout}}$, if we allow a variable size 
$d$ of the factors $\U$ and $\V$, we observe that 
$\Omega_{\scalebox{.6}{dropout}}$ naturally promotes for over-sized 
factorizations in the case of a fixed dropout rate $\theta$.
\item We show that the regularizer induced by dropout acts as a low-rank regularization strategy. Specifically, we show that if the dropout rate $\theta$ is chosen as a given function of $d$, then the optimization problem in \eqref{eq:dropoutMF} is related to the following matrix approximation problem
\begin{equation}
\min_{\A} \left[ \| \X - \A \|_F^2 + \gamma \| \A \|_\star^2 \right],
\end{equation}
where the squared nuclear norm is used to induce low-rank factorizations.
\item Furthermore, if we are given the global optimum factors $\U^\opt$ and 
$\V^\opt$ of  \eqref{eq:dropoutMF}, then $\A^\opt = (\U^\opt) 
(\V^\opt)^\top$ is the global optimum of \eqref{eq:MA} in the case of $\Xi(\A) = \| 
\A \|_\star^2$. Despite this result is derived in the case of variable size in the 
factorization, it is still applicable in the case of a fixed $d$.
\end{enumerate}

\myparagraph{Paper outline} In Section \ref{sez:rw} we briefly review the 
literature related to dropout. Sections 
\ref{sez:theory}, \ref{sez:nn} and \ref{sez:theory2} present our theoretical 
analysis, while numerical simulations are presented in Section \ref{sez:sim}. 
Concluding remarks are given in Section \ref{sez:end}. 

\section{RELATED WORK}\label{sez:rw}

There exists a broad and established literature which deals with either matrix 
factorization 
\eqref{eq:MF} or approximation \eqref{eq:MA}, also attempting to intertwine the 
two in either formal, algorithmic or applicative scenarios 
\cite{HintonRD,ADB,Bach:2008,Bach:2013,H14,H15,H17CVPR,H17}. Readers can
refer to \cite{Vidal:book} for a comprehensive dissertation. 

Orthogonally, in our work, we pursue a different perspective and we study dropout 
for MF \eqref{eq:dropoutMF}. To the best of our knowledge, apart from 
empirical validations in \cite{Zhai:CoRR15,He2016}, there is no theoretical 
analysis to understand which sort of regularization is implicitly performed by 
dropout on MF. In addition to solve this open problem, we discover that dropout can 
be used as a tool to interconnect matrix factorization \eqref{eq:MF} and 
approximation \eqref{eq:MA} problems. 

The origins of dropout can be traced back to the literature on learning 
representations from input data corrupted by noise 
\cite{Bishop:NC95,Bengio:ICML09,Rifai:CoRR11}. Since its original formulation 
\cite{DropoutCORR,DropoutJMLR}, many algorithmic variations have been 
proposed 
\cite{ImprovedDropout,FastDropoutRecurrent,DropConv,AdaptiveDropout,AnnealedDropout,2016arXiv161101353A,CurriculumDropout}.
Further, the empirical success of dropout for neural network training has 
motivated several works to investigate its formal properties from a theoretical point 
of view.
Wager et al. \cite{Wager:NIPS13} analyze dropout applied to the logistic loss for 
generalized linear models.  
Hembold and Long \cite{JMLR:v16:helmbold15a} discuss 
mathematical properties of the dropout regularizer (such as non-monotonicity and 
non-convexity) and derive a sufficient condition to guarantee a unique minimizer 
for the dropout criterion.
Baldi and Sadowski \cite{Baldi1,Baldi2} consider dropout applied to deep neural 
networks with sigmoid activations and prove that the weighted geometric mean of 
all of the sub-networks can be computed with a single forward pass.
Wager et al. \cite{NIPS2014_5502} investigate the impact of dropout on the 
generalization error in terms of the bias-variance trade-off. 
Gal and Ghahramani \cite{Gal2016Dropout} 
investigate the connections between dropout training and inference for deep 
Gaussian processes.

Many of these prior theoretical results required simplifying assumptions, and thus the results only hold in an approximate sense \cite{Wager:NIPS13,JMLR:v16:helmbold15a,Baldi1,Baldi2,Gal2016Dropout}. 
In contrast, we are able to characterize the regularizer induced 
by dropout for MF in an analytical manner which is still an open problem, actually 
motivated by the solid empirical performance scored by this paradigm 
\cite{Zhai:CoRR15,He2016}.

\section{DROPOUT FOR MATRIX FACTORIZATION}
\label{sez:theory} 

Given a fixed $m \times n$ matrix $\mathbf{X}$, we are interested in the problem 
of factorizing it as the product $\mathbf{U} \mathbf{V}^\top$, where 
$\mathbf{U}$ is $m \times d$ and $\mathbf{V}$ is $n \times d$, for some $d 
\geq \rho(\mathbf{X}) := {\rm rank}(\mathbf{X})$ that, in this Section, will be 
kept fixed for simplicity. In order to apply dropout to 
matrix factorization, we consider a random vector $\r = [r_1,\dots,r_d]$ whose 
elements are independently distributed as $r_i \sim {\rm 
Bernoulli}(\theta)$. 

\begin{remark}
	In what follows, to either avoid trivial cases or division by zero, we will assume $0 
	< \theta < 1$.  Let us stress that, our perspective is more general than currently 
	adopted practices for dropout training in neural networks where $\theta > 
	0.5$ (see \cite[Appendix A.4]{DropoutJMLR} for a list of typical values).
\end{remark}

By means of $\r$, we can apply dropout to the problem 
$\min_{\U,\V} \| \X - \U \V^\top \|_F^2$ as in \eqref{eq:dropoutMF}. To see why the 
minimization of \eqref{eq:dropoutMF} can be achieved by dropping 
out columns of $\U$ and $\V$, observe that if we use a gradient descent strategy, 
the gradient of the expected value is equal to the expected value of the gradient. 
Therefore, if we choose a stochastic gradient descent (SGD) approach in which the 
expected gradient at each iteration is replaced by the gradient for a fixed sample 
$\r$, we obtain that, while moving from $t$-th to $(t+1)$-th iteration,  the updated 
$\U^{t+1},\V^{t+1}$ factors are computed accordingly to Algorithm 
\ref{alg:drop_tr}. Thereby, the updates for the column of $\U^{t+1},\V^{t+1}$ 
are either performed or skipped accordingly to $\mathbf{r}^t$. In fact, at iteration 
$t$, the columns of $\U$ and $\V$ for which $r_i^t = 0$ are 
not updated, and the gradient update is only applied to the columns for which 
$r_i^t = 1$.  This observation precisely certifies that a SGD scheme\footnote{Note 
	that, when dropout training is applied in deep learning, the so-called 
	\emph{optimizer} (\eg, ADAM \cite{ADAM}) needs to be fixed a priori and 
	\emph{independently} with respect to the usage of dropout. Therefore, our 
	assumption of solving \eqref{eq:dropoutMF} with SGD is totally not-restrictive, 
	being furthermore in line with the current implementation practices that are 
	used 
	for training deep neural networks (see \cite{TF}).} applied to 
\eqref{eq:dropoutMF} is actually implementing dropout as originally proposed in 
\cite{DropoutCORR,DropoutJMLR}.
\begin{algorithm}[t!]
\For{$t = 1,2,\dots$}{Sample $\mathbf{r}^t$ elementwise from a $\Bt$. \\
 Compute the gradients \vspace{-10 pt}\begin{equation}
\begin{bmatrix} d\U^{t} \\ d\V^{t} \end{bmatrix} = \begin{bmatrix} 
(\X-\U^t\diag{\r^t}\V^{t\top})\V^t \\  (\X-\U^t\diag{\r^t}\V^{t\top})^\top\U^t 
\end{bmatrix}\vspace{-10 pt}\end{equation} with respect to $\U$ and $\V$, 
respectively. \\
Update the factors \vspace{-10 pt} \begin{equation}
\label{eq:SGD}
\begin{bmatrix} \U^{t+1} \\ \V^{t+1} \end{bmatrix} = \begin{bmatrix} \U^t \\ 
\V^t \end{bmatrix} + \frac{2\epsilon}{\theta}
\begin{bmatrix} d\U^t \\ d\V^t \end{bmatrix} \diag{\r^t},\vspace{-10 pt}
\end{equation}}
\caption{Dropout Training for MF}\label{alg:drop_tr}
\end{algorithm}

Corroborating the findings of various theoretical studies of dropout for general 
machine learning models
\cite{DropoutCORR,DropoutJMLR,JMLR:v16:helmbold15a,Baldi1,Baldi2,Gal2016Dropout,H17CVPR},
 we want to move tho the yet unexplored theory behind dropout for MF.
Namely, we are interested in proving that the latter
\eqref{eq:dropoutMF} is fully equivalent to a deterministic optimization problem of 
the form \eqref{eq:MF}, for a particular choice of $\Omega$. Ultimately, this will 
help us in better understanding of the implication of such random suppressions of 
columns that dropout is acting while the matrix $\X$ is factorized into $\U\V^\top$. 
This problem is tackled in the following theoretical result\footnote{\pmb{All proofs 
are in the Supplementary Material.}}.
\begin{thm}\label{thm:1}
	The two optimization problems \eqref{eq:MF} and \eqref{eq:dropoutMF} are 
	equivalent while choosing $\lambda$ and $\Omega$ in \eqref{eq:dropoutMF} to 
	be\vspace{-10pt}
	\begin{equation}\label{eq:Omd}
	\Omd(\U,\V) =  \sum_{k = 1}^d \| \mathbf{u}_k \|_2^2 \| 
	\mathbf{v}_k \|_2^2,\vspace{-5pt}
	\end{equation}
	where $\u_1,\dots,\u_d \in \mathbb{R}^m$ and $\v_1,\dots,\v_d \in 
	\mathbb{R}^n$ stand for the columns of $\U$ and $\V$ respectively and $
	\lambda = \tfrac{1 - \theta}{\theta}$.	
\end{thm}

Let us observe that, with the previous definition, $\lambda$ can takes all possible 
non-negative scalar values, since  as one can easily see, if we are interested in 
solving 
\eqref{eq:MF} with $\Omega = \Omd$ for a fixed $\bar\lambda$ 
value, we will always be able to find a fixed $\bar{\theta}$, $0 < \bar{\theta} < 
1$, such that $\bar\lambda = \tfrac{1 - \bar\theta}{\bar\theta}$. Indeed, since the 
relationship is invertible, one immediately gets $\bar\theta = \frac{1}{1 	
	+\bar{\lambda}}$.

The meaning of Theorem \ref{thm:1} is the following. Let consider the optimization 
problem \eqref{eq:MF} and fix $\Omega = \Omd$ as in \eqref{eq:Omd} and 
$\lambda = \frac{1 - \theta}{\theta}$. Then, the two optimization problems 
\eqref{eq:MF} and \eqref{eq:dropoutMF} are equivalent, where equivalence is 
intended in the strongest way possible, since, as proved in the Supplementary 
Material, for generic $\U$, $\V$, $d$ and $\theta$, we get
\begin{align}
&\mathbb{E}_{\mathbf{r}} \left\| \mathbf{X} - \dfrac{1}{\theta} 
\mathbf{U} {\rm diag}(\mathbf{r}) \mathbf{V}^\top \right\|_F^2 = \nonumber \\ 
&=  \| 
\mathbf{X} - 
\mathbf{U}  \mathbf{V}^\top \|_F^2 + \dfrac{1 - \theta}{\theta} \sum_{k = 
	1}^d \| \mathbf{u}_k \|_2^2 \| \mathbf{v}_k \|_2^2. \label{eq:superdooper}
\end{align}
The implications of \eqref{eq:superdooper} are clear: any stationary point of 
\eqref{eq:MF} with $\Omega = \Omd$ is also a stationary point of 
\eqref{eq:dropoutMF} and vice versa. Furthermore, 
the two problems have the same global minimum since, despite the non convexity of the optimization problem, in the case of MF, there exist some theoretical guarantees to ensure the existence of a global minimizer due to the fact that the regularizer is shaped as product of columns of the factors \cite{Recht:2010,H14,H15,H17}. For instance, while building on ideas derived from convex relaxations, general frameworks such as \cite{H15} allow for the analysis of non-convex factorizations and derives sufficient conditions for optimality condition of the non-convex optimization problem.

%
%
%

In this work, we characterize the optimum of droput with MF with a closed-form matrix approximation problem with squared nuclear norm regularization.

\section{CONNECTIONS WITH THE NUCLEAR NORM}\label{sez:nn}

For $\A \in \mathbb{R}^{m \times n}$, its nuclear norm, also termed the trace 
norm or Schatten-Von Neumann $1$-norm,  
\begin{equation}\label{eq:nn}
\| \A \|_\star = \sum_{i = 1}^{\min(m,n)} \sigma_i(\A)
\end{equation}
is defined as the sum of its singular values 
$\sigma_i(\A)$, $i = 1,\dots,\min(m,n)$. Within many machine 
learning problems 
\cite{yuan2007dimension,Argyriou2008,Candes2009,cabral2011matrix,Olsen}, 
the usage of \eqref{eq:nn} is motivated by the fact that $\| \A 
\|_\star$ is a convex relaxation for the rank $\rho(\A)$ of $\A$. Indeed, it is proved 
that the underlying low rank solution 
can be recovered by minimizing \eqref{eq:nn} under certain conditions 
\cite{candes2010power,Recht:2010}.

In order to establish a connection between \eqref{eq:nn} and the regularizer 
\eqref{eq:Omd}, let us consider the following result.
\begin{thm}[\emph{Variational form of the nuclear norm}]
	\begin{equation}\label{eq:var_f_nn} \vspace{-10pt}
	\| \X \|_\star = \inf_{\U,\V \colon \U\V^\top = \X} \sum_{k =1}^{d} \| \u_k \|_2 \| 
	\v_k \|_2.
	\end{equation}	
\end{thm}
We can find a close similarity between computing the infimum of \eqref{eq:Omd} 
over $\U$ and $\V$ such that $\U\V^\top = \X$ and \eqref{eq:var_f_nn}, except to 
a point. Instead of summing the product Euclidean norms $\| 
\cdot \|_2$ among the columns of $\U$ and $\V$ as in \eqref{eq:var_f_nn}, in 
$\Omega_{\scalebox{.6}{dropout}}$, we are summing the products of \emph{squared} 
Euclidean norms $\| \cdot \|_2^2$ among the columns of $\U$ and $\V$. Although 
this difference may seem marginal, this is not actually the case.

\begin{remark}\label{remark_warning}
	Let fix two arbitrary random matrices $\U$ and $\V$ of sizes $m \times 
	d$ 
	and 
	$n \times d$, respectively. Now, consider the case of a variable size of factorization $d$. Then,
	\begin{align}
	0 = \inf_{d,\U,\V} \sum_{k = 1}^d \| \mathbf{u}_k \|_2^2 \| 
	\mathbf{v}_k \|_2^2 
	\quad \text{s.t.} \, \begin{cases}
	d \geq \rho( \X) \\
	\U\V^\top = \X
	\end{cases}
	\end{align}
	since we can observe  that
	\begin{equation}\label{eq:warning}
	\Omega_{\scalebox{.6}{dropout}}\left(\hspace{-.05cm}\hspace{-.05cm} \tfrac{\sqrt{2}}{2}[\mathbf{U},\hspace{-.05cm} \mathbf{U}], 
	\tfrac{\sqrt{2}}{2}[\mathbf{V},\hspace{-.05cm} \mathbf{V}] \hspace{-.05cm}\right) \vspace{-.05cm}\hspace{-.05cm} = \hspace{-.05cm}\hspace{-.05cm} \frac{1}{2} 
	\Omega_{\scalebox{.6}{dropout}}\left( \mathbf{U},\hspace{-.05cm} \mathbf{V} \right).
	\end{equation}
	So if we minimize the objective function \eqref{eq:dropoutMF} - or, equivalently, 
	 \eqref{eq:MF} with $\Omega = \Omd$ - over $\U$,$\V$ and $d$ 
	as well, 
we may trivially lower the value of the objective function through Algorithm 
\ref{alg:casini} which, clearly does not promote $\U\V^\top$ to be close to $\X$ in 
any case.
\begin{algorithm}[t!]
Randomly initialize $\U^0$ and $\V^0$ \;
\For{ $t = 1,2,\dots$}{Update the factors\vspace{-10pt}\begin{equation}
\U^{t +1} = \tfrac{\sqrt{2}}{2} [\U^t, \U^t], \V^{t + 1} =
\tfrac{\sqrt{2}}{2} [\V^t, \V^t] \label{eq:crazy_opt}\vspace{-10pt}
\end{equation}}
\caption{Pathological oversizing in the factors}\label{alg:casini} 
\end{algorithm}
\end{remark}


In the previous observation, we analyzed what happens if we relax $d$ from a fixed 
and (heuristically) chosen value to be one of the active variables of the 
optimization. The latter aspect is actively investigates as a research topic 
\cite{Recht:2010,H14,H15,H17,H17CVPR} and many algorithms have been 
proposed with this respect so that, in our case, we can take advantage of any of 
those when asked to optimize
\begin{equation}\label{eq:dropoutMFd}
	\min_{\U,\V,d} \mathbb{E}_{\mathbf{r}} \left\| \mathbf{X} - \tfrac{1}{\theta} 
	\mathbf{U} {\rm diag}(\mathbf{r}) \mathbf{V}^\top \right\|_F^2	\tag{3$'$}
\end{equation}
over $d$ as well. In the present work, we will not investigate this aspect, since 
it's not primarily related to our scope. Differently, we allow $d$ to be variable for the 
sake of improving the theoretical understanding dropout of MF. Through this 
modification, additionally, we bridge the gap between dropout for MF 
\eqref{eq:dropoutMF}, its equivalent reformulation \eqref{eq:MF} with 
$\Omega= \Omd$ and the matrix factorization problem \eqref{eq:MA} where 
$\Xi(\A) = \| \A \|_\star^2$. 




\section{VARIABLE SIZE FACTORS}\label{sez:theory2}

In this Section, we want to establish a connection between the class of problems 
\eqref{eq:MA} and dropout for MF, as explained in the previous Section can be 
formulated either as \eqref{eq:dropoutMF} or as its fully deterministic counterpart 
\eqref{eq:MF} with $\Omega$ as in \eqref{eq:Omd}.

In order to fill such gap, we are interested in observing whether 
there exists a way to choose $\theta$ to depend upon the size of the factorization 
$d$, such that we can avoid the pathological optimization scheme \ref{alg:casini} 
which promotes over-sized factorizations.

\begin{prop}\label{prop:prop}
	For a given $p$, $0 < p < 1$, define
	\begin{equation}\label{eq:theta_d}
	\theta(d) = \dfrac{p}{d -  (d -1)p}
	\end{equation}
	where $d$ refers to the size of the factorization for $\X$, quantified 
	in terms of columns of $\U$ and $\V$. Then
	\begin{align}
	\tfrac{1 - \theta(2d)}{\theta(2d)}\Omega_{\scalebox{.6}{dropout}}\hspace{-.1 cm}\left( \hspace{-.1 cm}
	\tfrac{\sqrt{2}}{2}[\mathbf{U},\hspace{-.05 cm} \mathbf{U}], 
	\tfrac{\sqrt{2}}{2}[\mathbf{V},\hspace{-.05 cm} \mathbf{V}] \hspace{-.1 cm} \right)\hspace{-.1 cm} = \hspace{-.1 cm}\tfrac{1 - 
	\theta(d)}{\theta(d)}
	\Omega_{\scalebox{.6}{dropout}}\left(\hspace{-.05 cm} \mathbf{U}, 
	\mathbf{V}\hspace{-.05 cm} \right)\hspace{-.1 cm}. \nonumber
	\end{align}
\end{prop}


In Proposition \ref{prop:prop}, we modify the dropout retain probability $\theta$ to 
be function of $d$, while also depending on a novel hyper-parameter $p$. We will 
discuss later on the meaning and the necessity of introducing it, but for now, let's 
say that $p$ is fixed in the range $]0,1[$. 

In principle, the only guarantee that Proposition \ref{prop:prop} ensures is that 
the choice $\theta = \theta(d)$ as in \ref{eq:theta_d} prevents the over-sizing in 
the factorization. Indeed, other issues may arise and, potentially, one may be asked 
to change $\theta(d)$ in order to accommodate for them. Actually, we can show 
that the definition \eqref{eq:theta_d} is able to solve \emph{all} the problematics 
of dropout applied to MF with variable size due to the following result.

\begin{prop}\label{prop:LowConvEnv}
	For $\theta = \theta(d)$ as defined in \eqref{eq:theta_d}, $\tfrac{1-p}{p} \| 
	\X\|^2_\star$ is the lower convex envelope\footnote{One defines lower convex 
	envelope of a function $f$ as the supremum over all convex functions $g$ such 
	that $g \leq f$.} of $$\Lambda(\mathbf{X}) = 
	\inf_{\U,\V,d} \tfrac{1 - \theta(d)}{\theta(d)}  \Omd(\U,\V)$$ subject to $d \geq \rho(\X)$ and $\U\V^\top = \X$.
\end{prop}


\begin{figure*}[t!]
	\includegraphics[width=\textwidth,keepaspectratio]{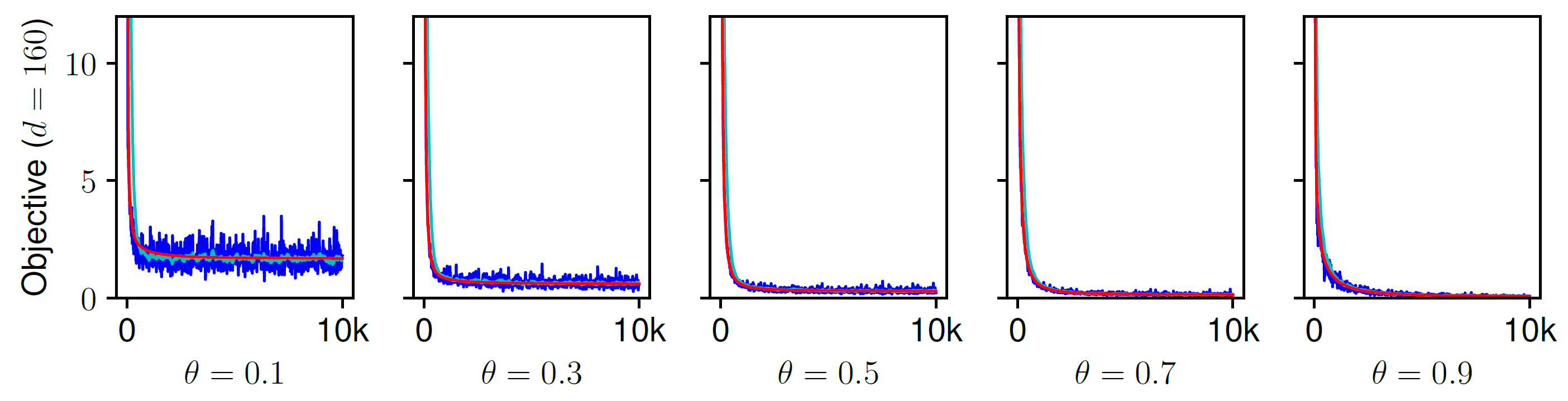}
	\caption{For $\theta \in \{0.1, 0.3, 0.5, 0.7, 0.9\}$ and $d = 160$ 
		we compare dropout for MF \eqref{eq:dropoutMF} (blue) and its deterministic counterpart (red). The exponential moving average of the stochastic objective is in cyan. Best viewed in color.}
	\label{fig:obj_curves}
\end{figure*}

Let us remember that, as we show in Remark \ref{remark_warning}, when we 
compute the infimum of $\Omd(\U,\V)$ over $\U,\V,d$ such that $d \geq \rho(\X)$ 
and $\U\V^\top = \X$, we get zero \underline{if} the dropout retain probability 
$\theta$ is fixed. Differently, when $\theta = \theta(d)$ is allowed to be a function 
of $d$ as in \eqref{eq:theta_d}, we immediately get that the infimum of $\tfrac{1 - 
\theta(d)}{\theta(d)} \inf_{\U,\V,d} \Omd(\U,\V)$ is \underline{not} zero and, 
ancillary, this prevents pathological scheme like \eqref{eq:crazy_opt} to decrease 
the objective value of \eqref{eq:dropoutMFd} without really approximating $\X$. 
Differently, Proposition \ref{prop:LowConvEnv} guarantees that the adaptation of 
the dropout rate $\theta$ is able to constrain the regularizer in terms of a convex 
lower bound for it, the lower convex bound being (a scaled version) of the squared 
nuclear norm $\| \X \|_\star^2$. This enables us to retrieve a 
stronger connection\footnote{Let us clarify that such connection is not totally 
unexpected, 	even in the variable size case, since as proved in the Supplementary 
Material, the variational form \eqref{eq:var_f_nn} holds when we optimize over $d$ 
	in addition to $\U$ and $\V$.} between dropout regularizer and (squared) nuclear 
	norm, achieving a disciplined linkage between the two.

Actually, taking advantage of Proposition \ref{prop:LowConvEnv}, we can provide a 
stronger theoretical result, which, on the one hand, establishes a direct connection 
between dropout for MF with variable size and squared nuclear norm regularization.

\begin{thm}\label{thm:ciaobelli}
	Let $\U^\opt$ and $\V^\opt$ the $m \times d^\opt$ and $n \times d^\opt$ 
	optimal factors that achieves the global optimum of dropout for MF 
	\eqref{eq:dropoutMFd} with $\theta = \theta(d)$ as in \eqref{eq:theta_d} for 
	some fixed hyper-parameter $p$, $0 < p < 1$. Then $\A^\opt = 
	(\U^\opt)\cdot(\V^\opt)^\top$ is the global minimizer of 
	\begin{equation}\label{eq:CLB}
	\min_{\A} \left[ \| \X - \A \|_F^2 + \dfrac{1 - p}{p} \| \A \|_\star^2 \right],
	\end{equation}
	which corresponds to optimizing over $\A \in \mathbb{R}^{m \times n}$ the problem \eqref{eq:MA} with $\Xi = \| \cdot \|_\star^2$ and 
	$\gamma = \tfrac{1 - p}{p}$.
\end{thm}


Theorem \ref{thm:ciaobelli} achieves our targeted goal of exploiting dropout as a 
leap between matrix factorization \eqref{eq:MF} and approximation \eqref{eq:MA} 
problems. As we did in Section \ref{sez:theory}, thanks to the marginalization 
through expectation as in \eqref{eq:dropoutMFd}, we are able to condensate all the 
stochastic suppression of 
columns in the factors into a fully deterministic problem \eqref{eq:MF} with 
$\Omega = \Omd$, and, also, the same equivalence holds when $d$ is variable.
Actually, the real reason to do that is, in such a case, we can define a variable 
dropout retain probability $\theta = \theta(d)$ as in \eqref{eq:theta_d} and retrieve 
that dropout for MF is equivalent to the optimization problem \eqref{eq:CLB}. 
Precisely, that ``equivalence'' should be interpreted as follows: the global optimum 
$(\U^\opt,\V^\opt)$ of \eqref{eq:dropoutMFd} provides for free the global optimum 
$\A^\opt = (\U^\opt)\cdot(\V^\opt)^\top$ for \eqref{eq:CLB}.

Equation \eqref{eq:bard} is useful also to understand the role of the 
hyper-parameter $p$ that was introduced within the definition of 
\eqref{eq:theta_d}. In fact, 
the necessity of the dependence on $p$ in $\theta(d)$ \eqref{eq:theta_d} is 
dictated from the exigence of allowing a variable regulation for the squared nuclear 
norm regularization \eqref{eq:CLB}. In fact, consistently with our goal of using 
dropout as a leap in between matrix factorization \eqref{eq:MF} and approximation 
\eqref{eq:MA}, by defining the dropout retain probability $\theta$, we are able, on 
the one hand, to find $\lambda$ in \eqref{eq:MF} as $\lambda = \tfrac{1 - 
\theta}{\theta}$ and, on the other hand, when $\theta(d) = \tfrac{p}{d - (d - 1 
)p}$, we select $\gamma$ in \eqref{eq:MA} to be $\gamma = \tfrac{1 - p}{p}$. 
Let us observe that having dropout retain probability that depends upon 
hyper-parameters has been already proposed in the literature (\eg 
\cite{CurriculumDropout}).

As a final remark, since the objective function of \eqref{eq:CLB} is strictly 
convex, the existence and uniqueness of the global minimizer of \eqref{eq:CLB} is 
guaranteed and, moreover, it can be expressed through the following closed form 
solution.

\begin{thm}\label{thm:Aopt}
Let $\X = \mathbf{L} \boldsymbol{\Sigma} \mathbf{R}^\top$ be the singular 
valued decomposition of $\X$. The optimal solution $\A^\opt$ to \eqref{eq:CLB} 
is given by 
\begin{equation}\label{eq:Aopt}
\A^\opt = \mathbf{L} \S_{\mu}(\boldsymbol{\Sigma}) 
\mathbf{R}^\top
\end{equation}
where $\S_{\mu}(\sigma) = \max(\sigma - \mu, 0)$ defines the shrinkage thresholding operator\footnote{For a general scalar $x$, one usually defines $\S_{\mu}(x) = {\rm sgn}(x)\max(|x| - \mu, 0)$, but, here, due to the non-negativity of the singular values $\sigma > 0$, we will exploit the simplified expression $\S_{\mu}(\sigma) = \max(\sigma - \mu, 0)$.} \cite{Vidal:book} applied entrywise to the singular values $\sigma_i(\X)$ of $\X$ and
\begin{equation}\label{eq:mu}
\mu = \frac{1 - p}{p + (1 - p)\bar{d}} \sum_{i = 1}^{\bar{d}} \sigma_i(\X)
\end{equation}
where $\bar{d}$ denotes the largest integer such that
\begin{equation}\label{eq:bard}
\sigma_{\bar{d}} (\X) > \frac{1 - p}{p + (1 - p)\bar{d}} \sum_{i = 1}^{\bar{d}} \sigma_i(\X).
\end{equation}
\end{thm}


The convex lower bound \eqref{eq:CLB} to dropout for MF allows a 
closed-form solution in terms of the singular value decomposition of $\X$. While 
keeping the same singular vectors, the singular values are instead
massaged by means of the shrinkage thresholding operator $\S_\mu$ where 
$\mu$ is data dependent. Moreover, in order to compute it, on needs to found 
$\bar d$ as in \eqref{eq:bard} before computing \eqref{eq:Aopt}. 

We can interpret the latter points as follows: dropout for MF with variable size is 
sort of acting a dimensionality reduction technique, which is very close to PCA 
\cite{Vidal:book}. However, two differences arise: first, the number of 
principal components is not (heuristically) fixed but dropout learns it to be 
$d^\opt = \bar{d}$. Second, the top $\bar{d}$ singular values are not directly
used for the projection, but, instead, we shrink them in a way 
that is adaptively induced by the data itself. Since we find this connection between 
dropout for MF and the sort of adaptive PCA described below, we can ultimately 
state that the following. Dropping out columns in the factors acts as a regularizer 
which promotes spectral sparsity for low-rank solutions.

\begin{figure*}[t!]
	\centering
	\includegraphics[keepaspectratio, width=0.19\textwidth]{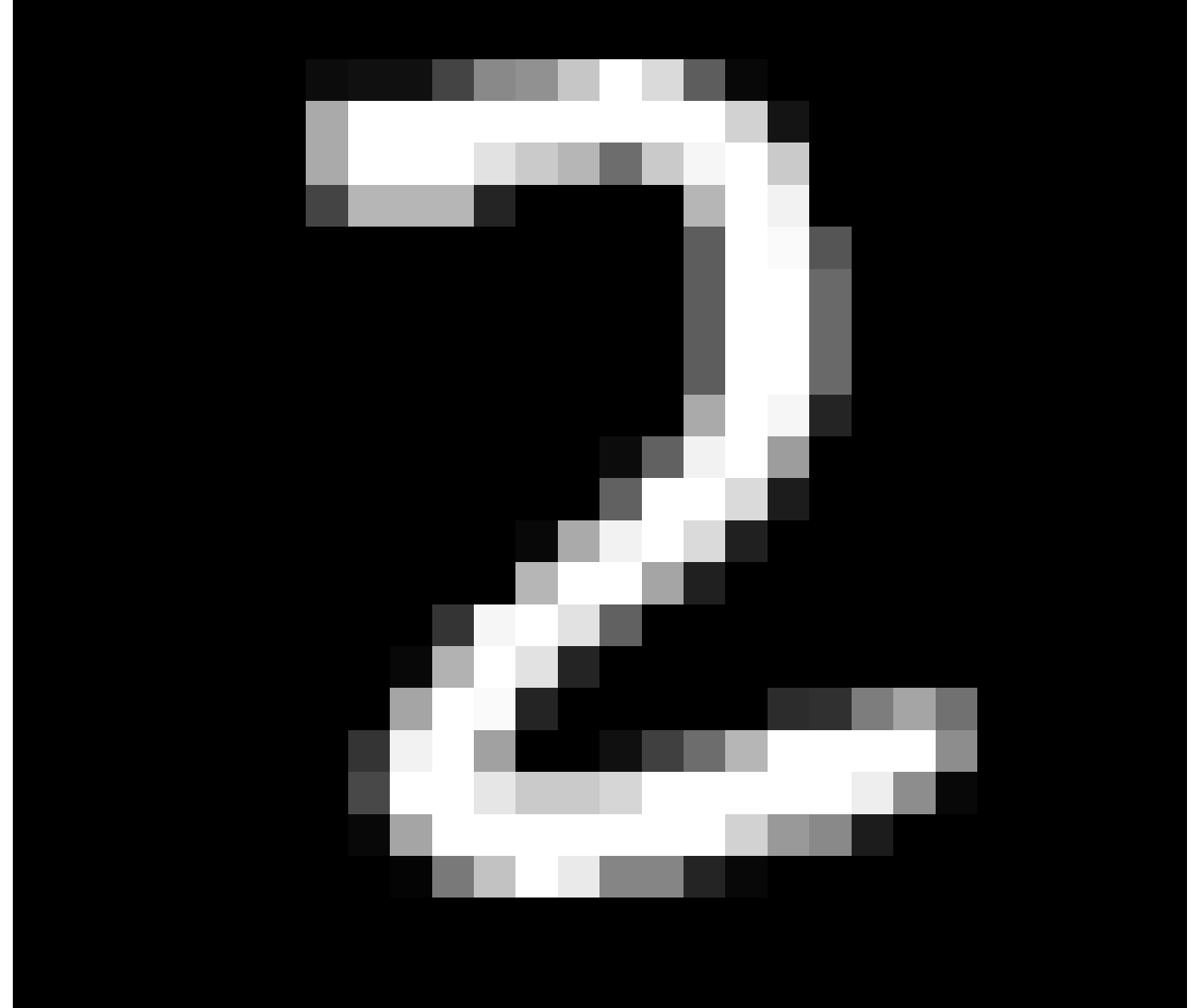}\quad
	\includegraphics[keepaspectratio, width=0.19\textwidth]{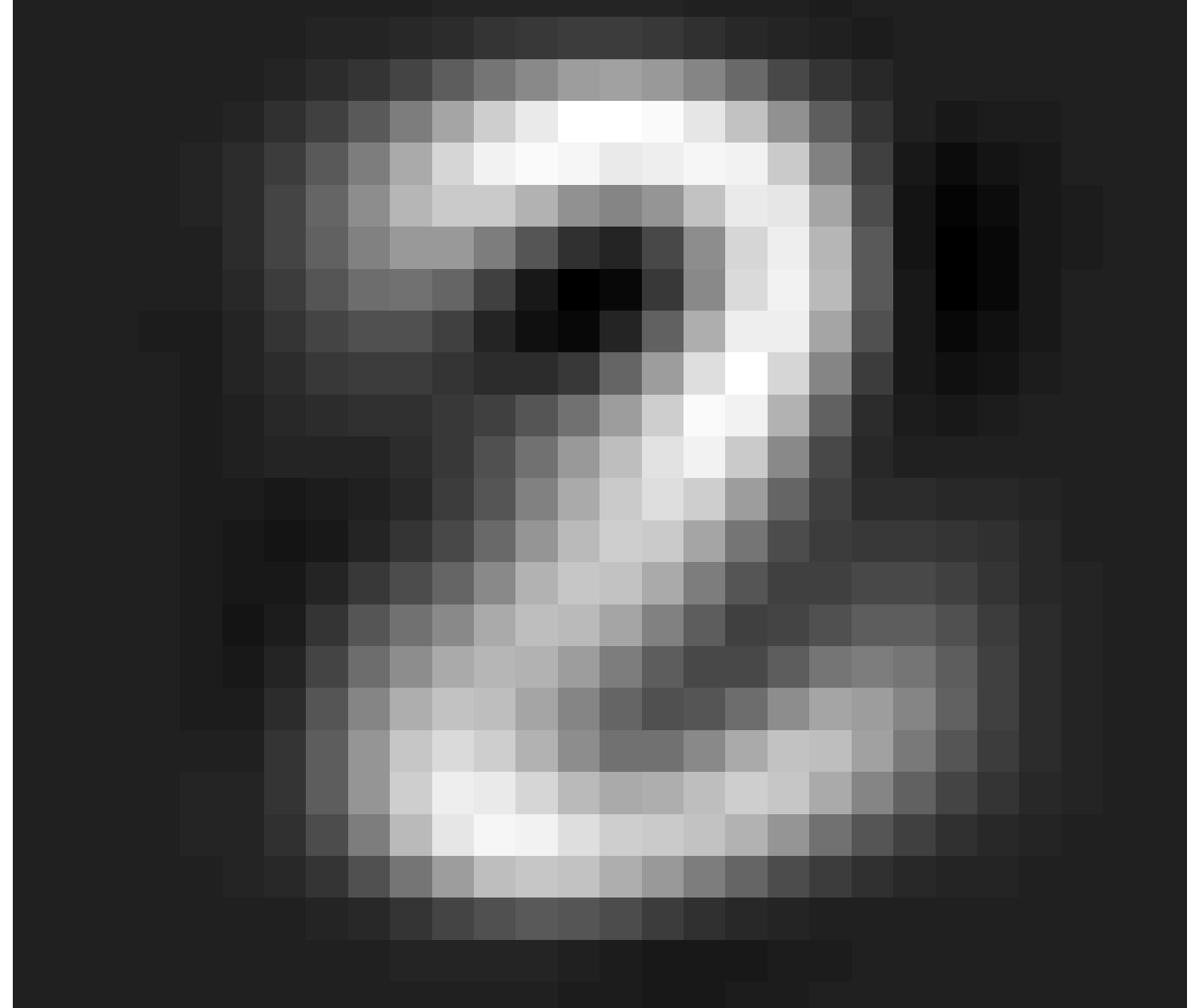}%
	\includegraphics[keepaspectratio, width=0.19\textwidth]{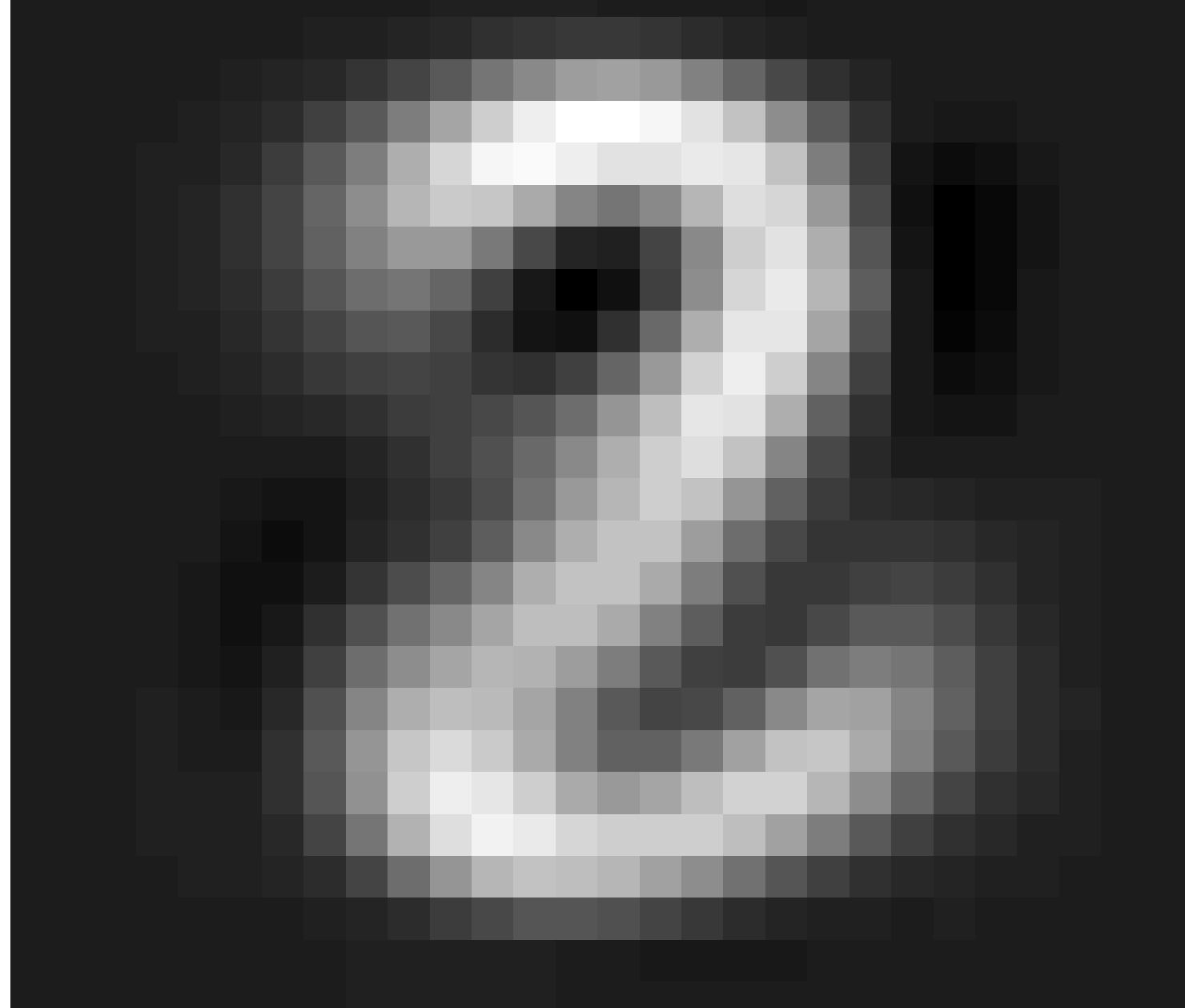}\quad
	\includegraphics[keepaspectratio, width=0.19\textwidth]{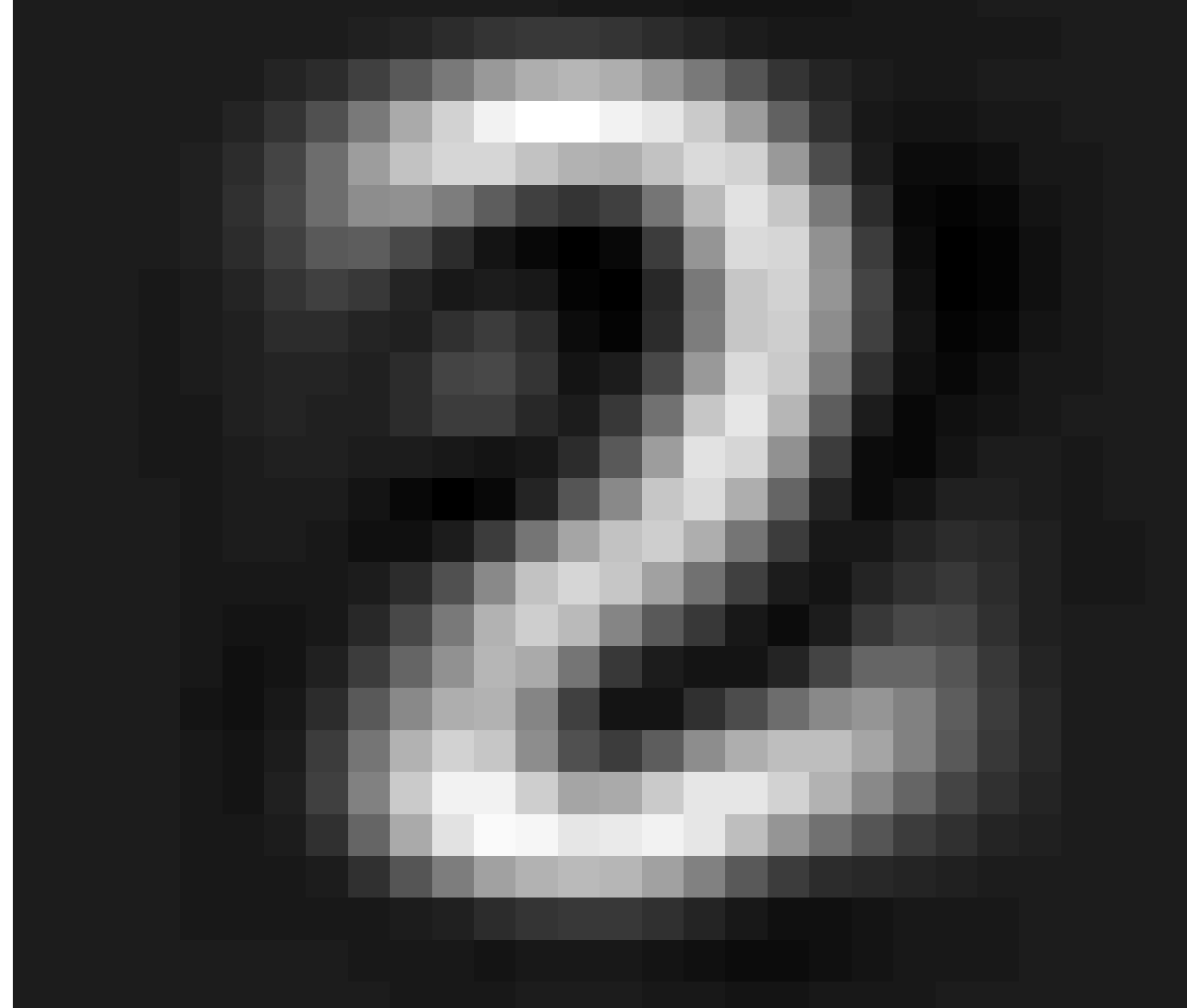}%
	\includegraphics[keepaspectratio, width=0.19\textwidth]{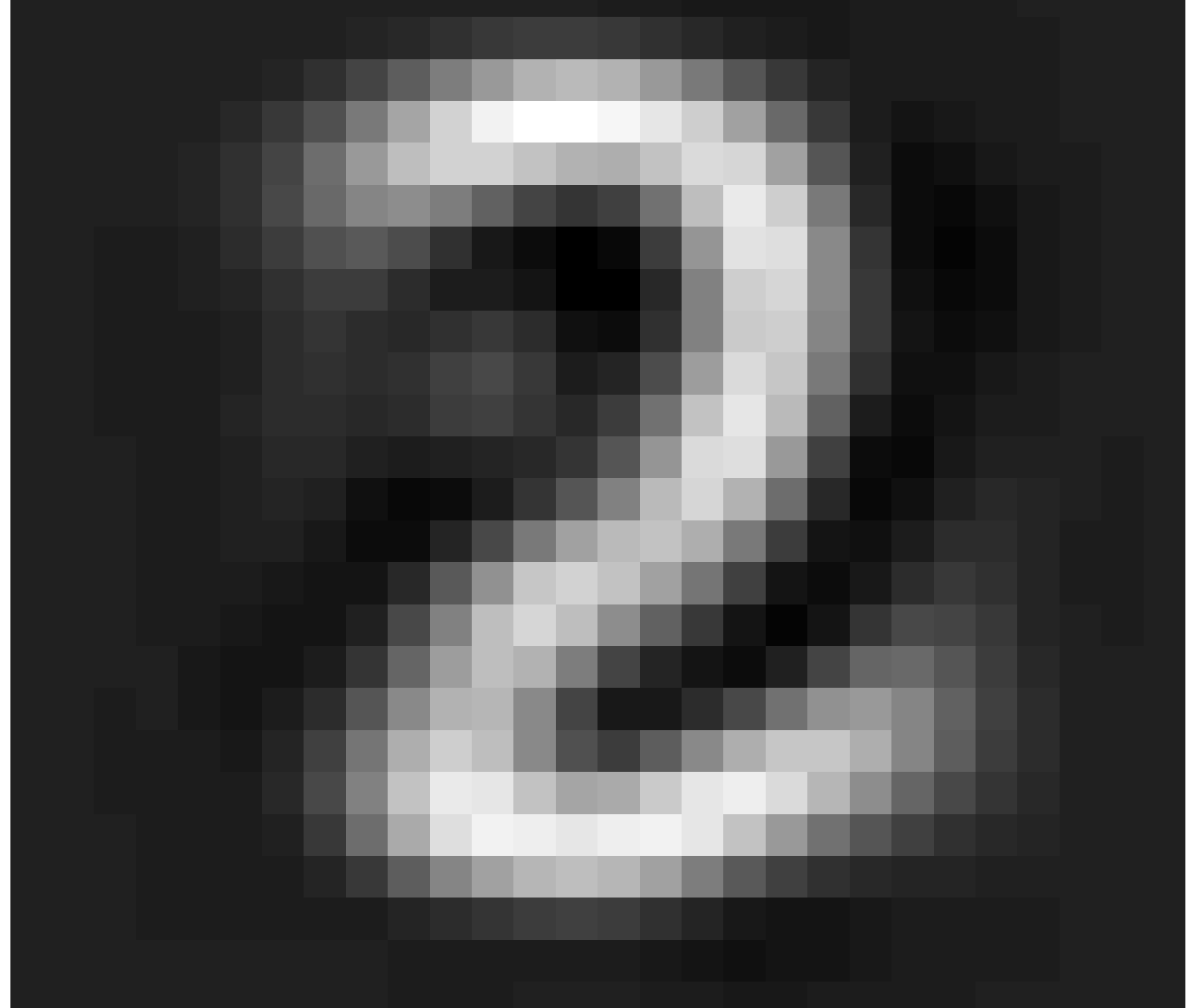}\\
	\includegraphics[keepaspectratio, width=0.19\textwidth]{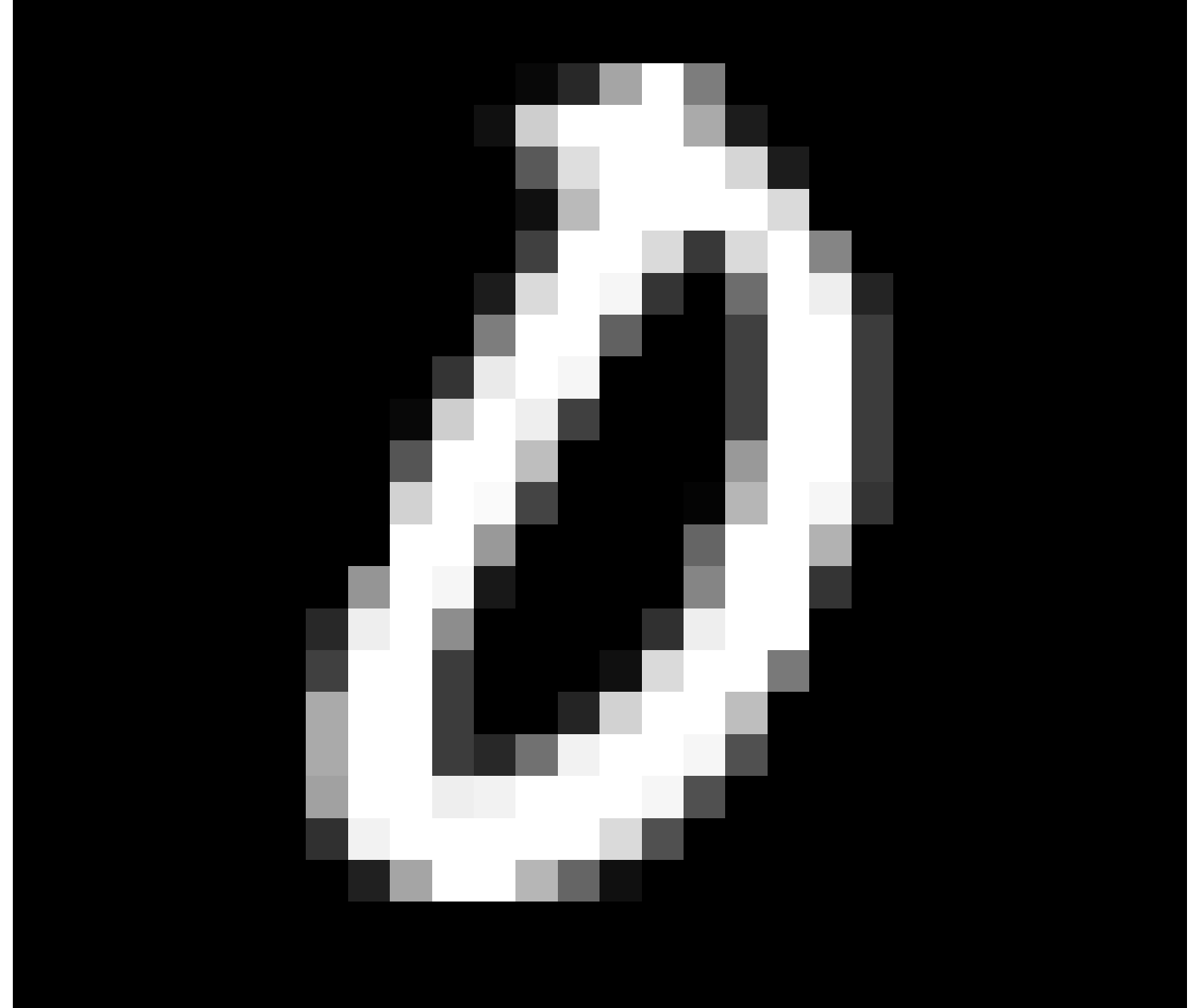}\quad
	\includegraphics[keepaspectratio, width=0.19\textwidth]{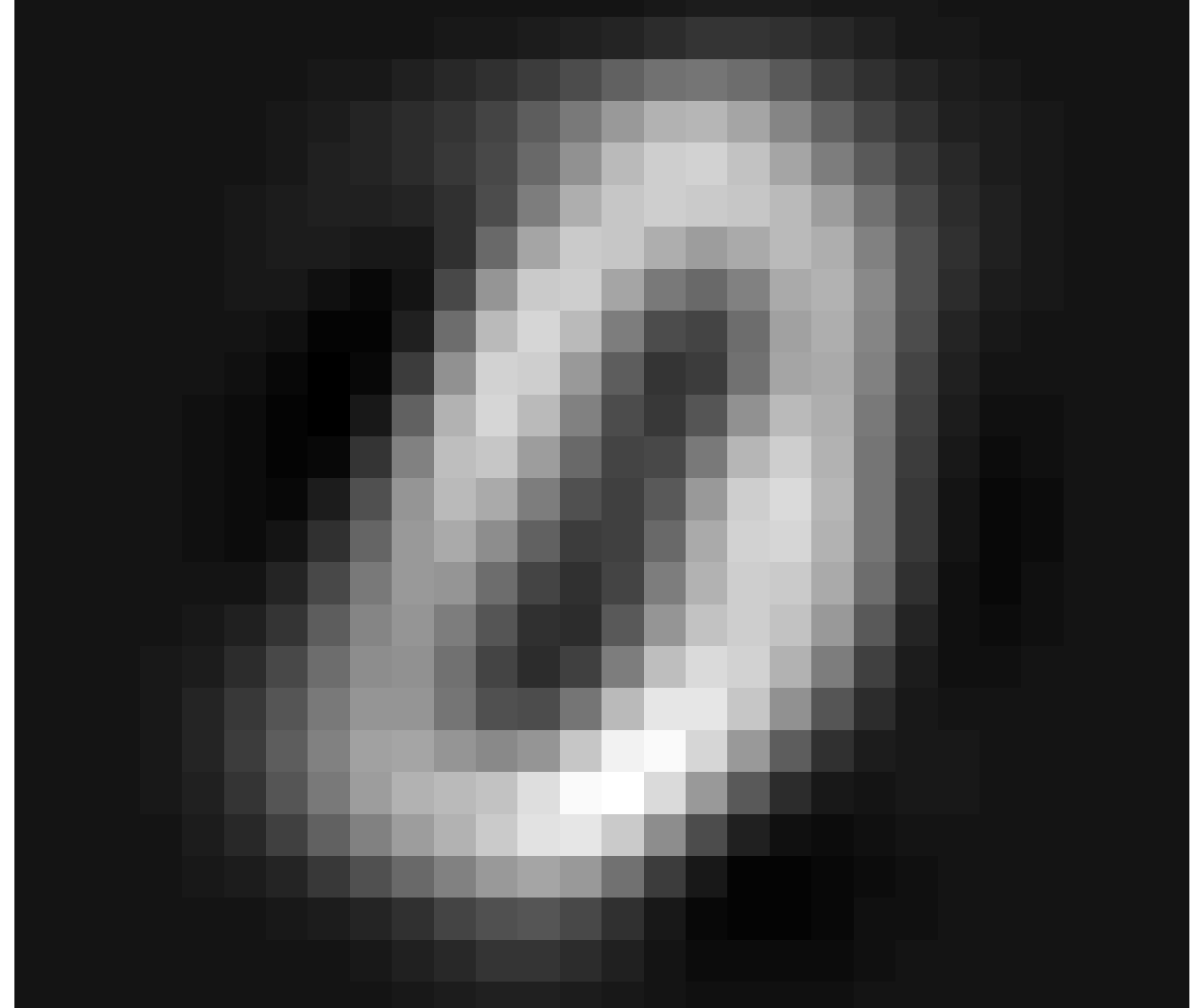}%
	\includegraphics[keepaspectratio, width=0.19\textwidth]{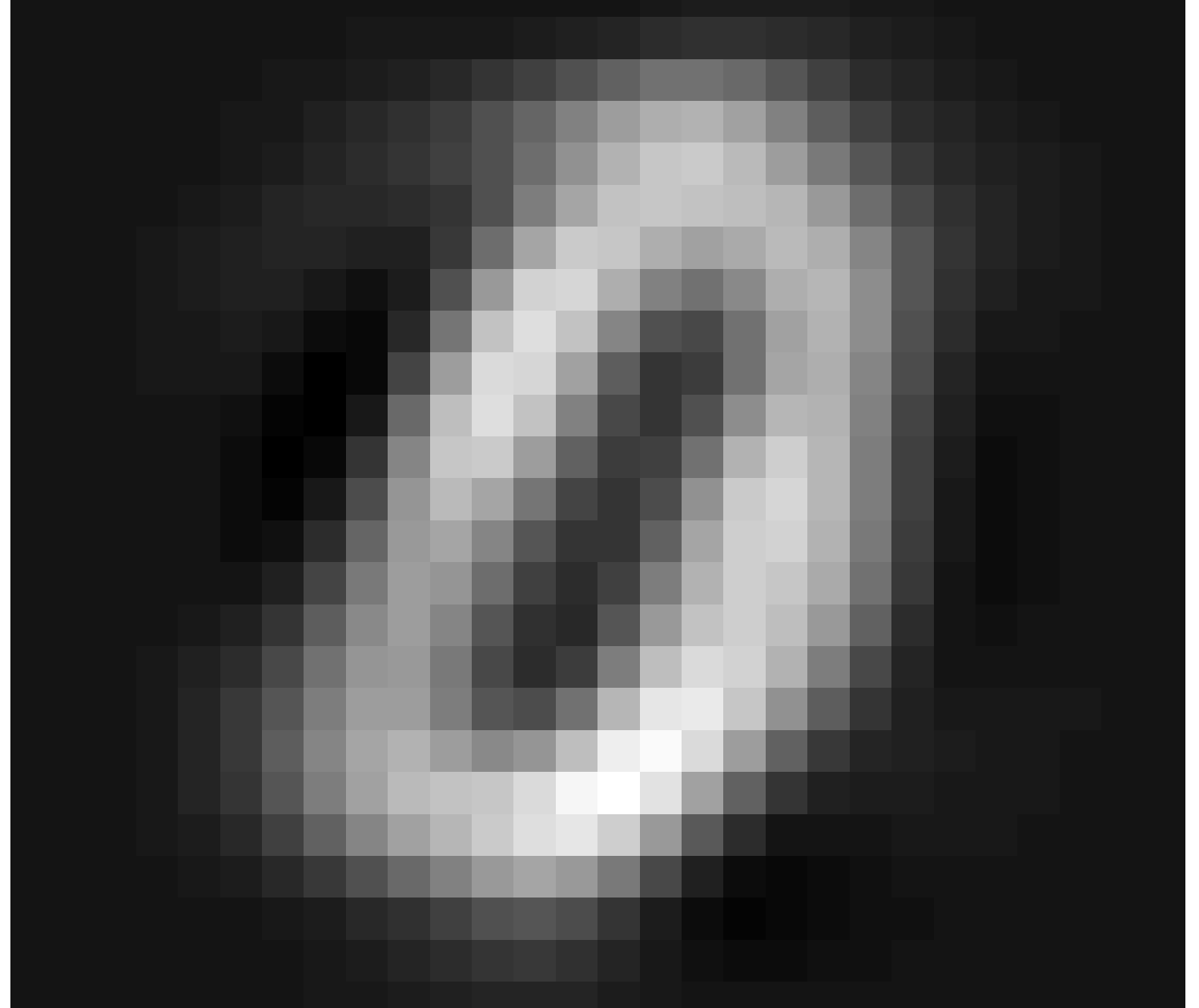}\quad
	\includegraphics[keepaspectratio, width=0.19\textwidth]{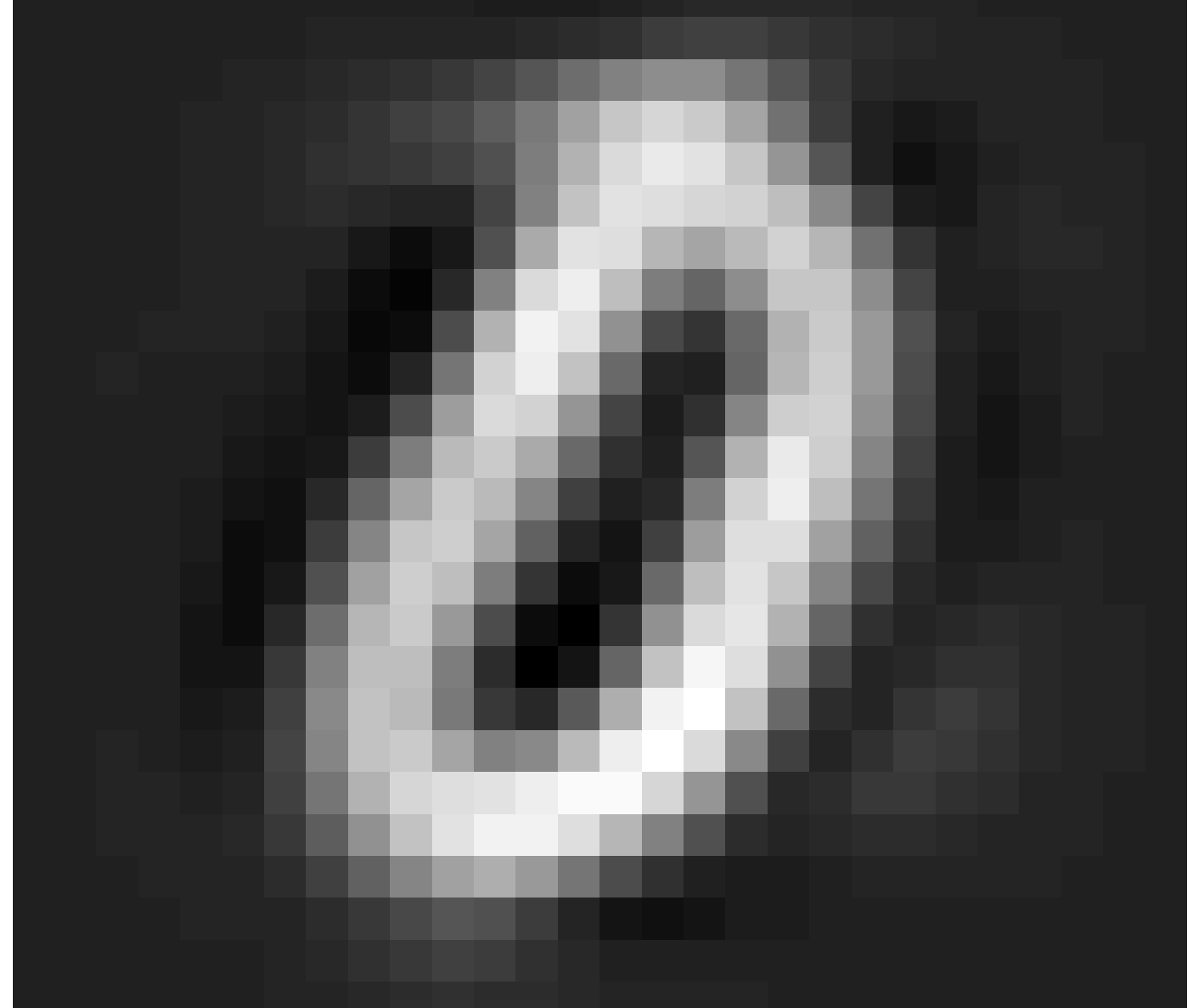}%
	\includegraphics[keepaspectratio, width=0.19\textwidth]{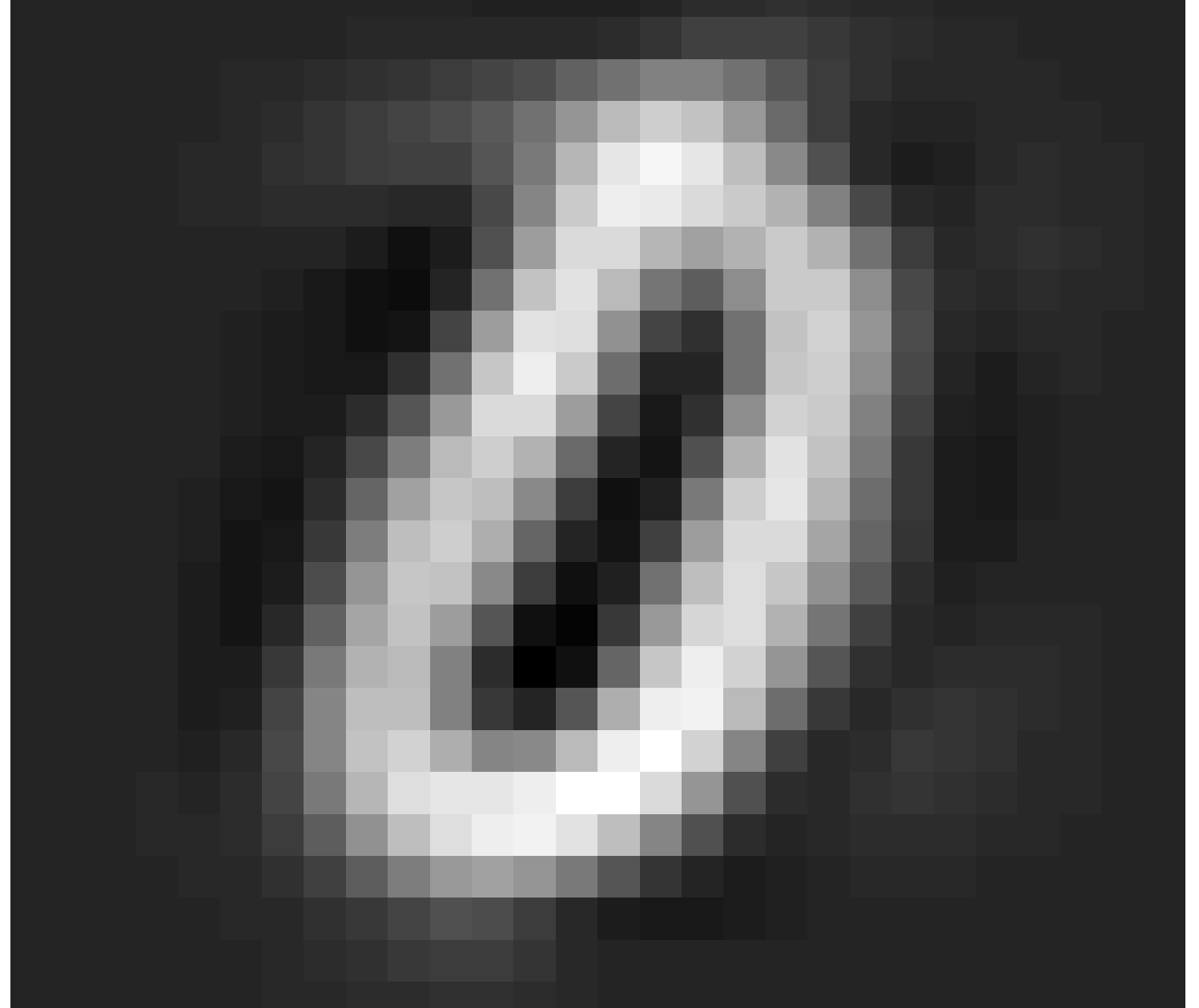}\\
	\includegraphics[keepaspectratio, width=0.19\textwidth]{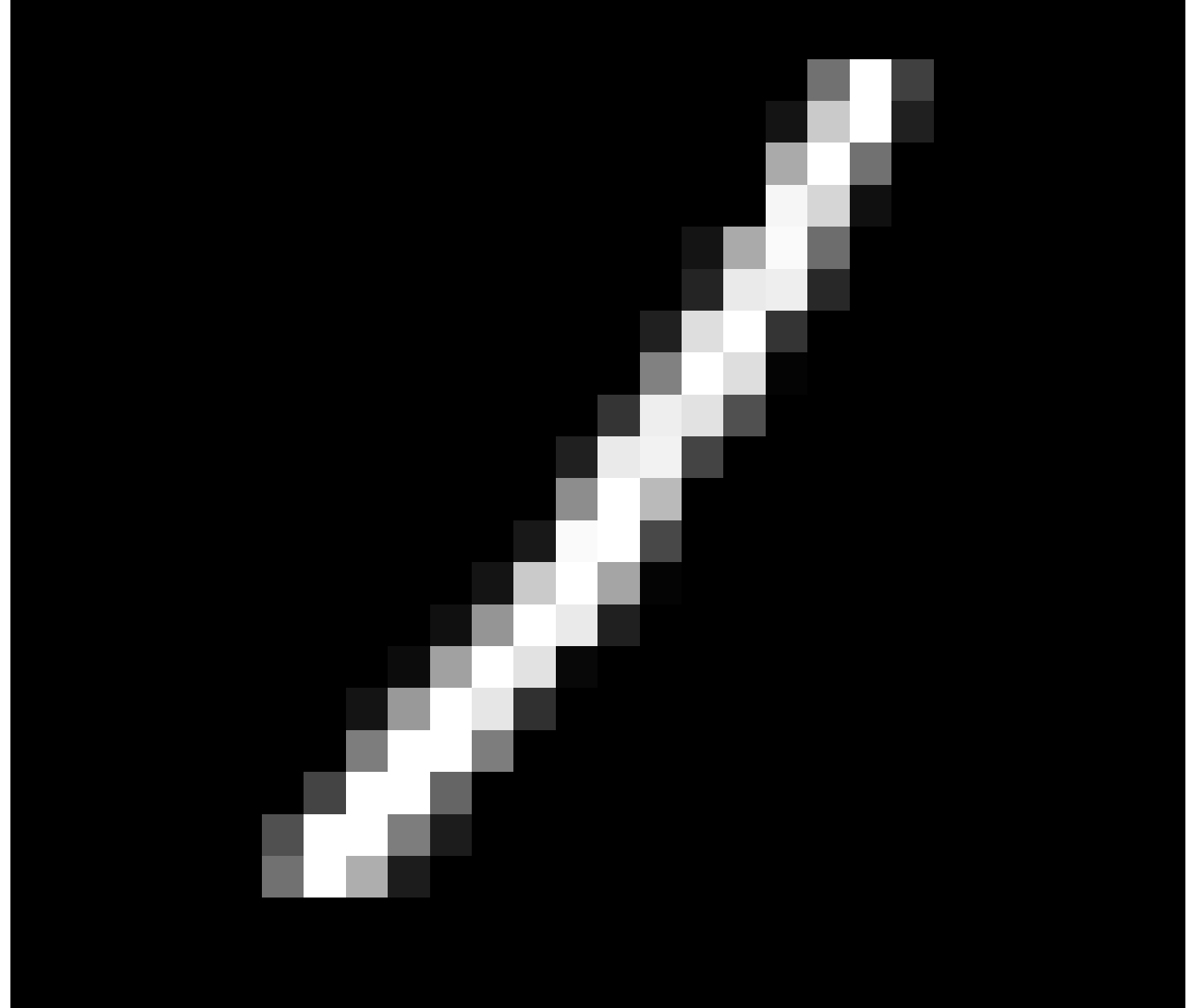}\quad
	\includegraphics[keepaspectratio, width=0.19\textwidth]{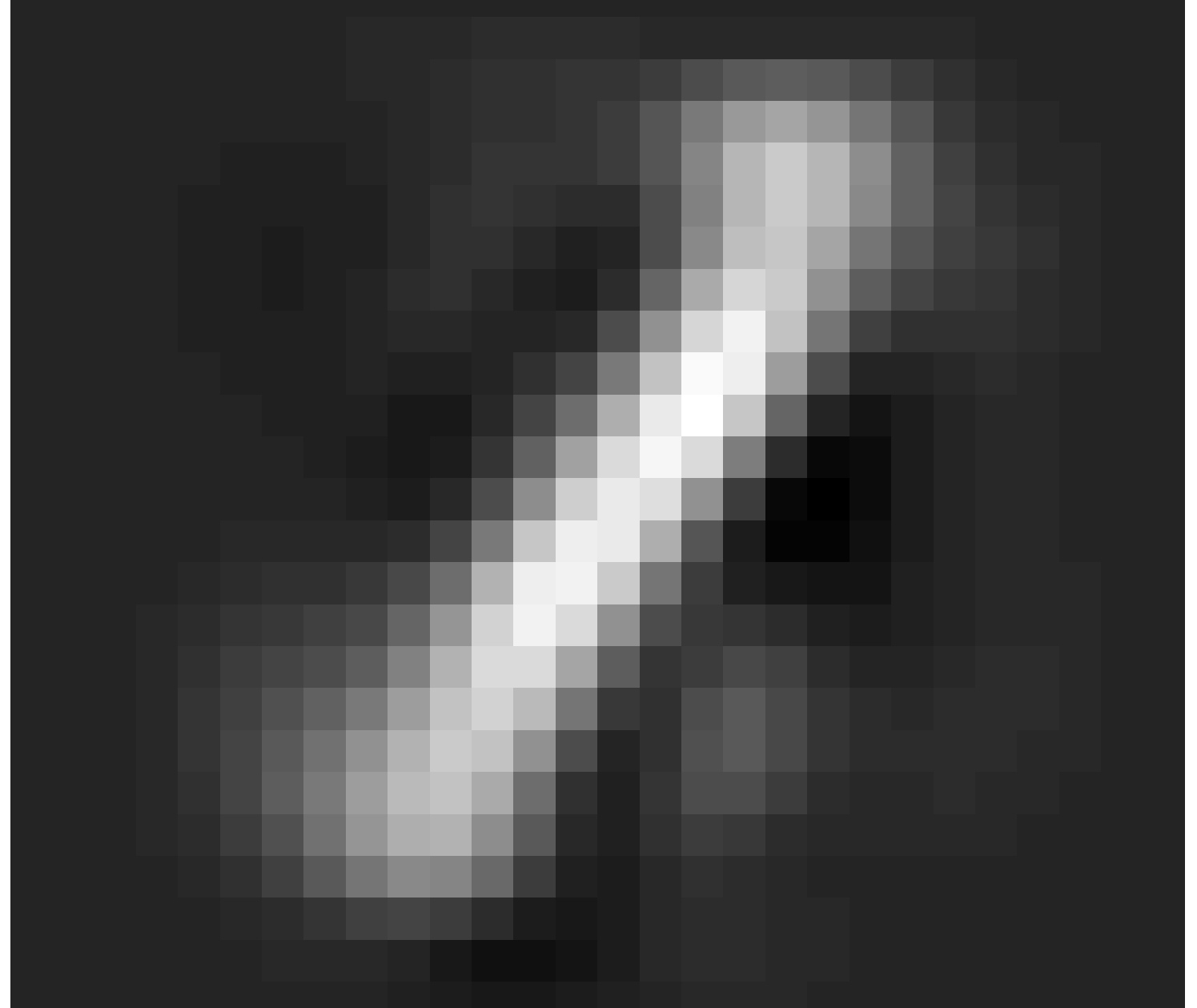}%
	\includegraphics[keepaspectratio, width=0.19\textwidth]{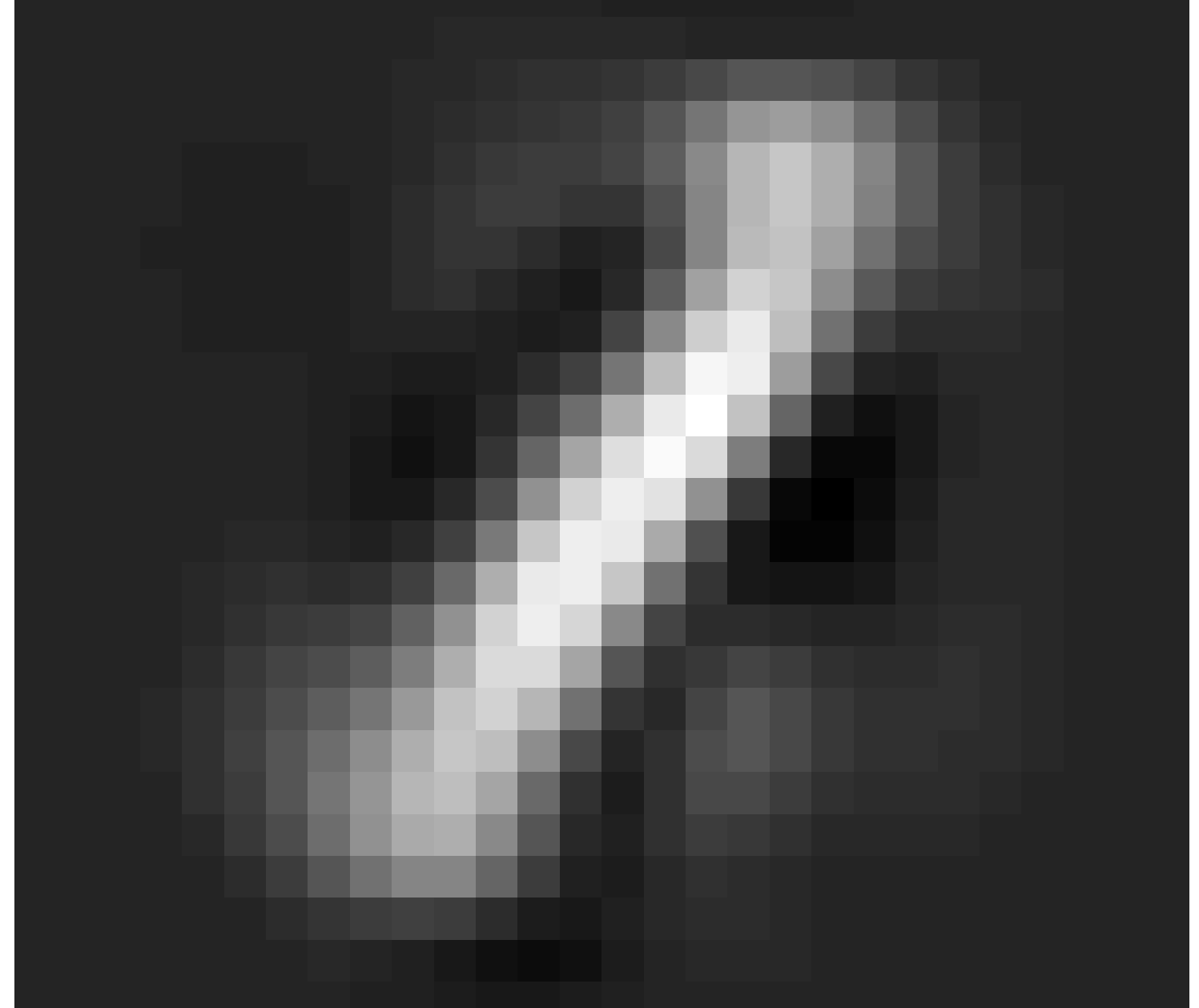}\quad
	\includegraphics[keepaspectratio, width=0.19\textwidth]{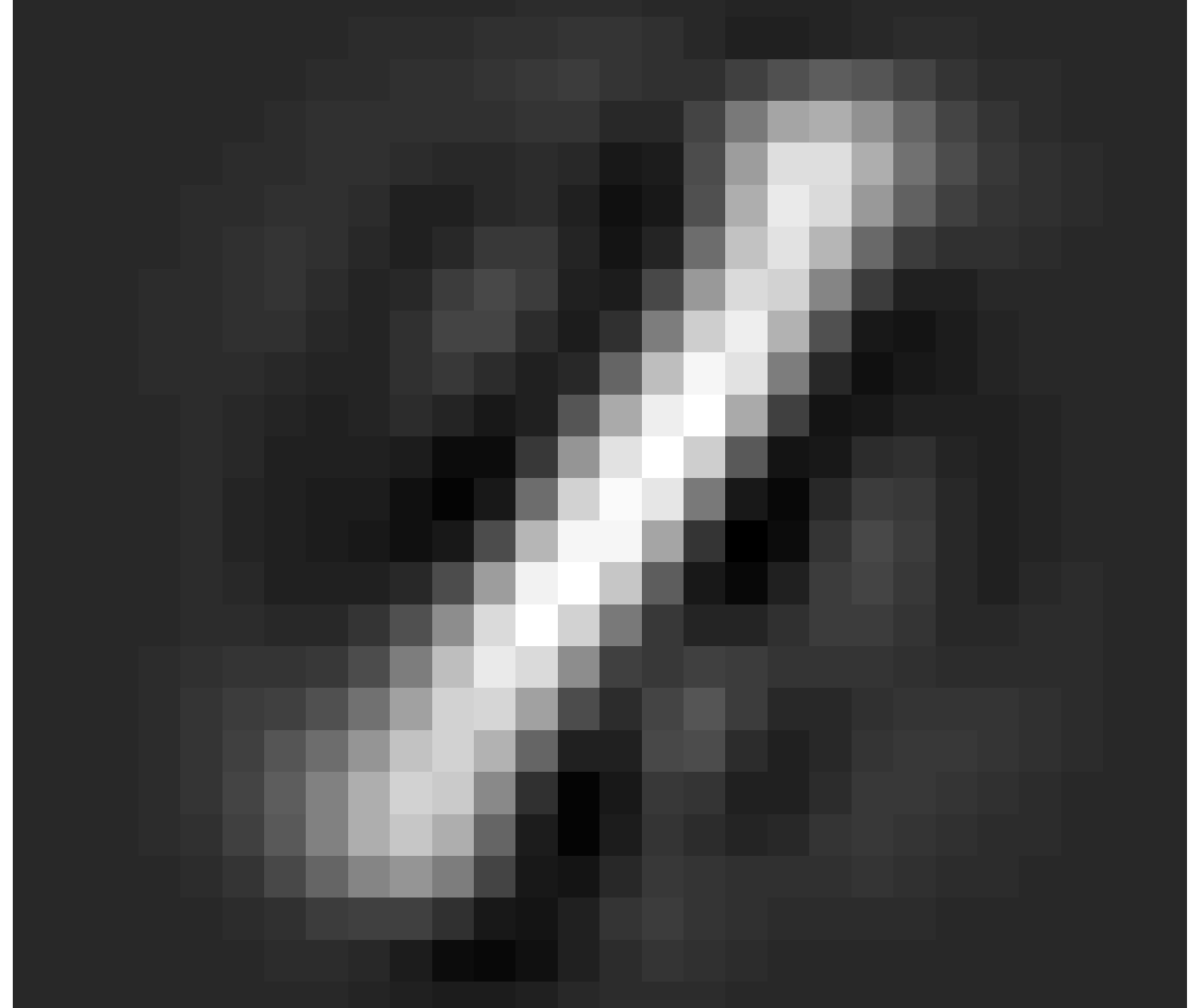}%
	\includegraphics[keepaspectratio, width=0.19\textwidth]{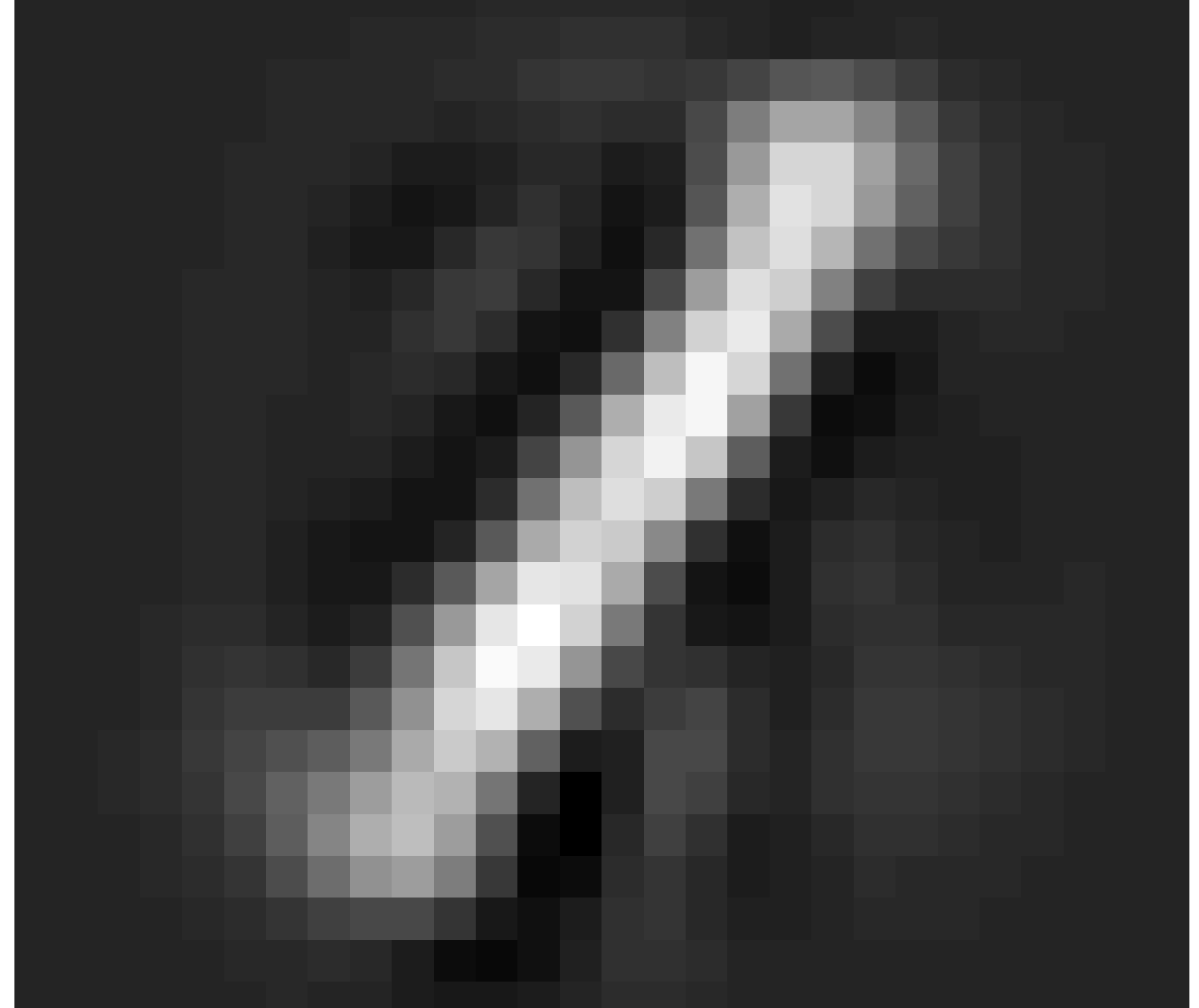}\\
	\includegraphics[keepaspectratio, width=0.19\textwidth]{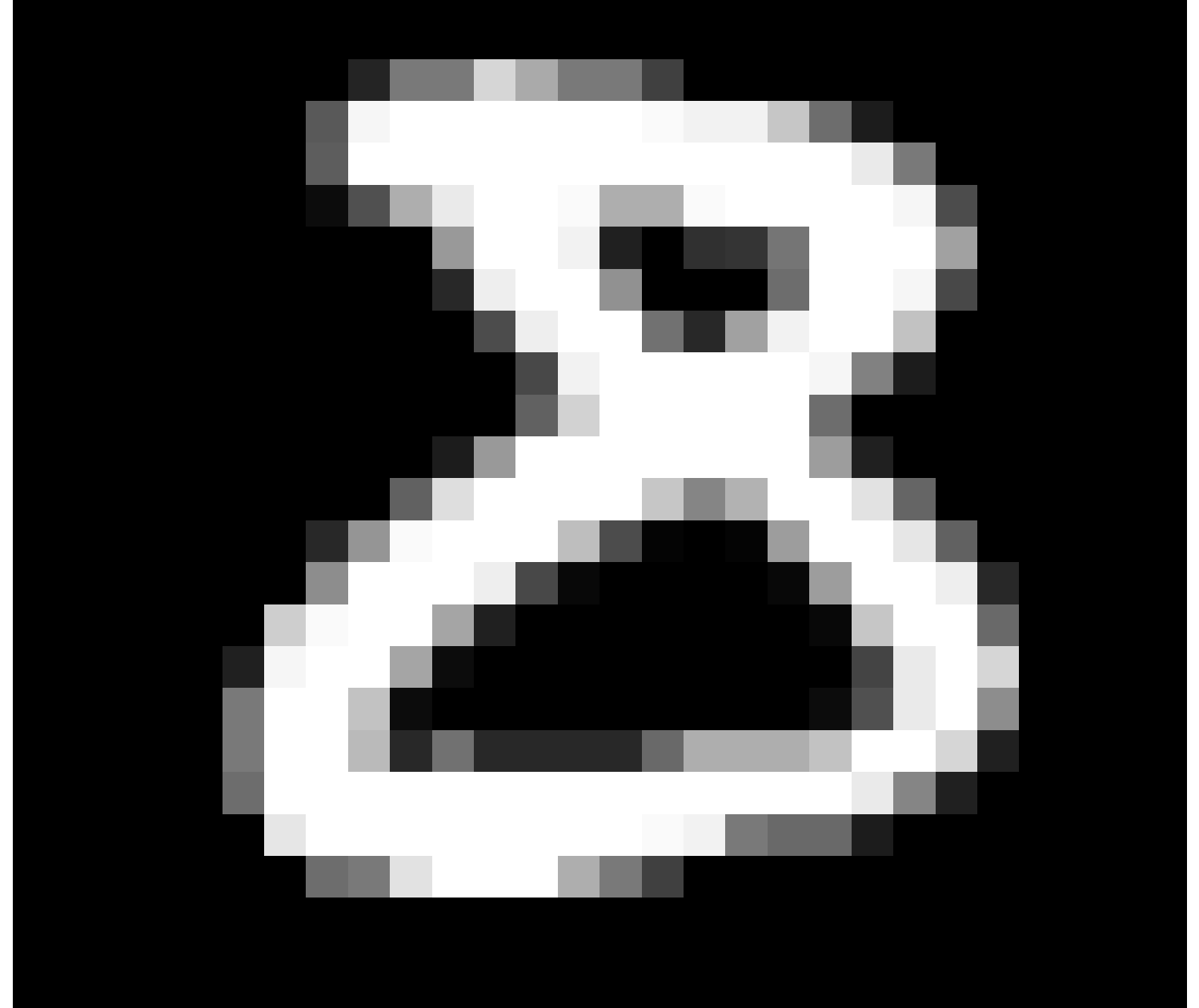}\quad
	\includegraphics[keepaspectratio, width=0.19\textwidth]{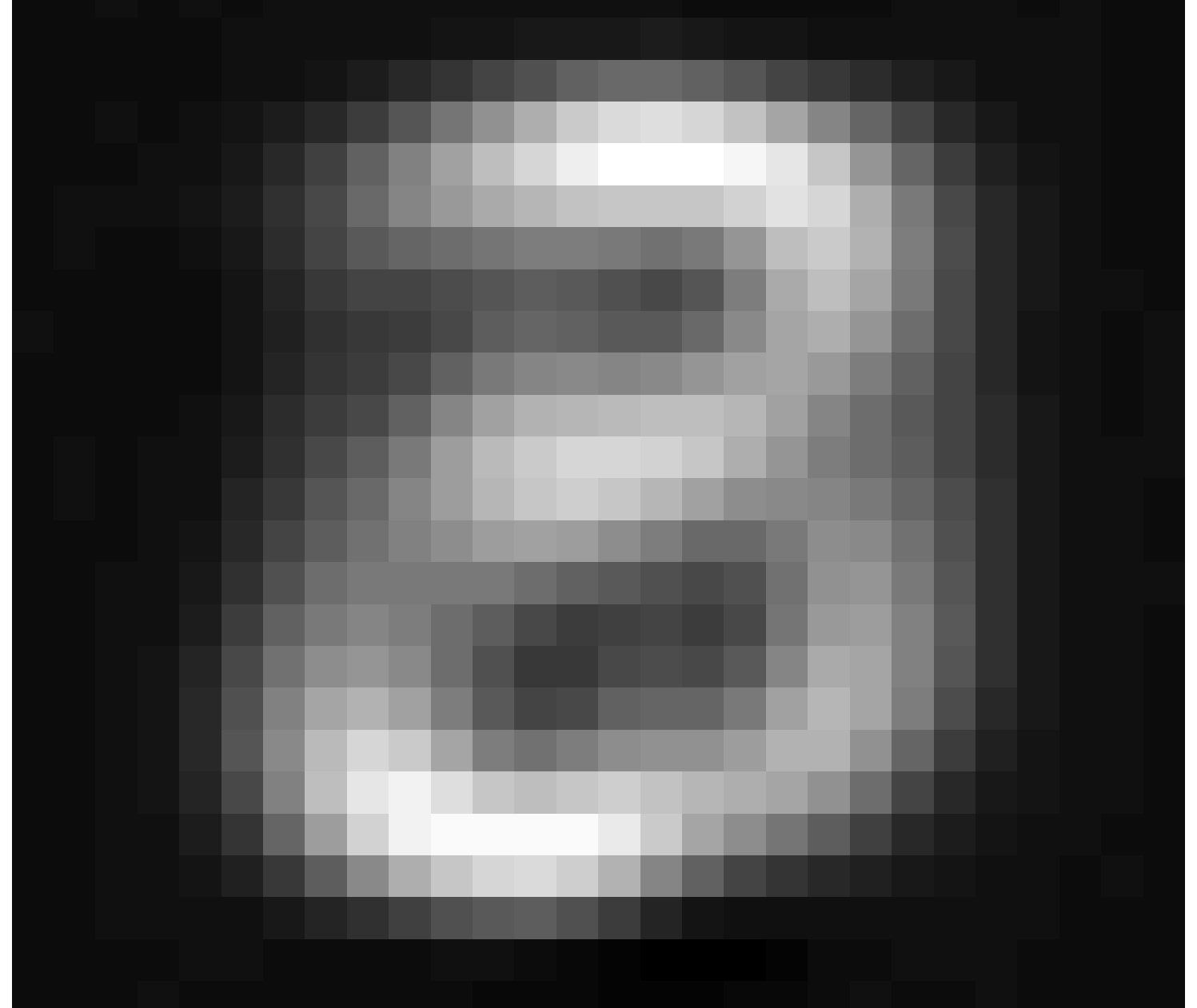}%
	\includegraphics[keepaspectratio, width=0.19\textwidth]{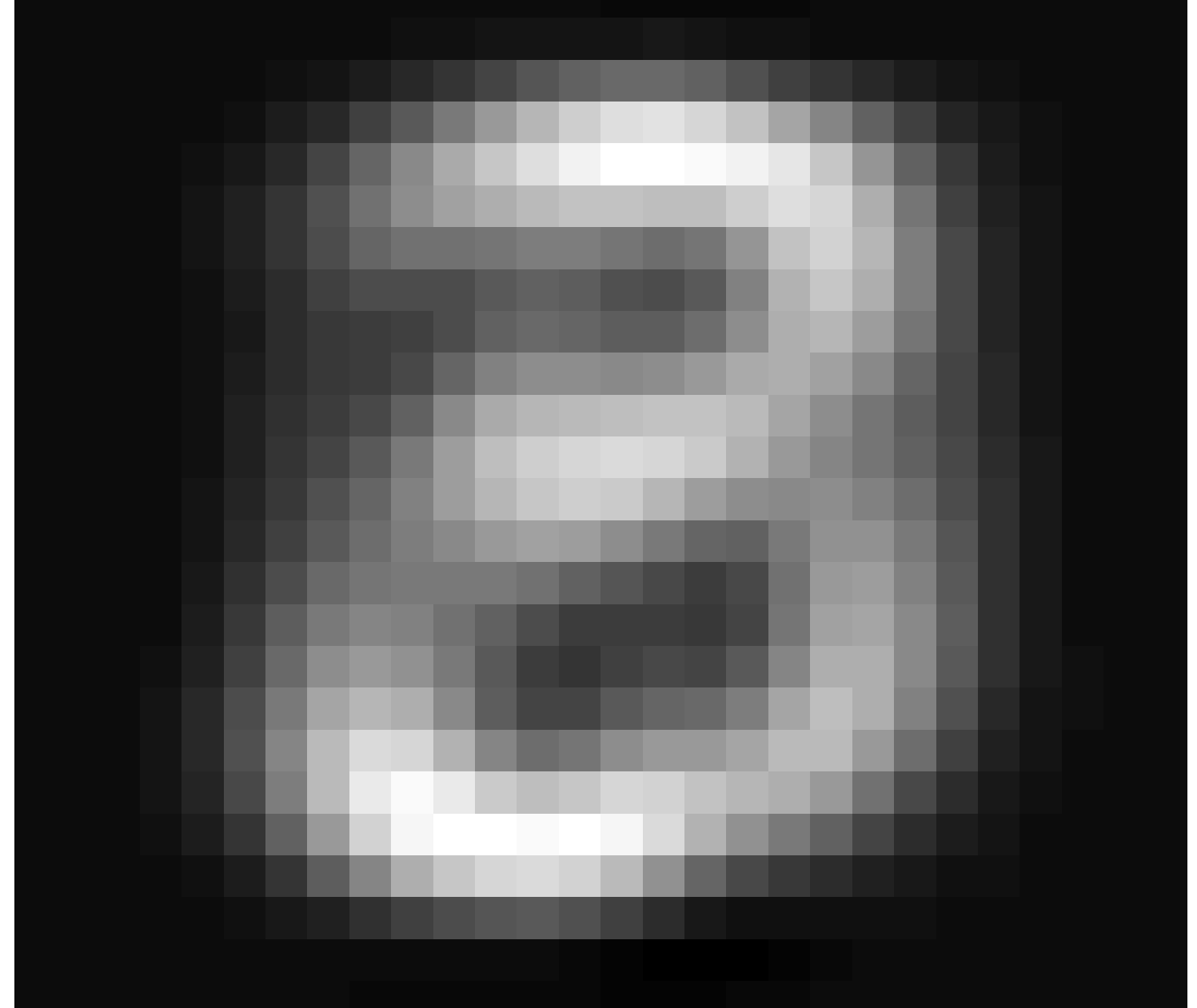}\quad
	\includegraphics[keepaspectratio, width=0.19\textwidth]{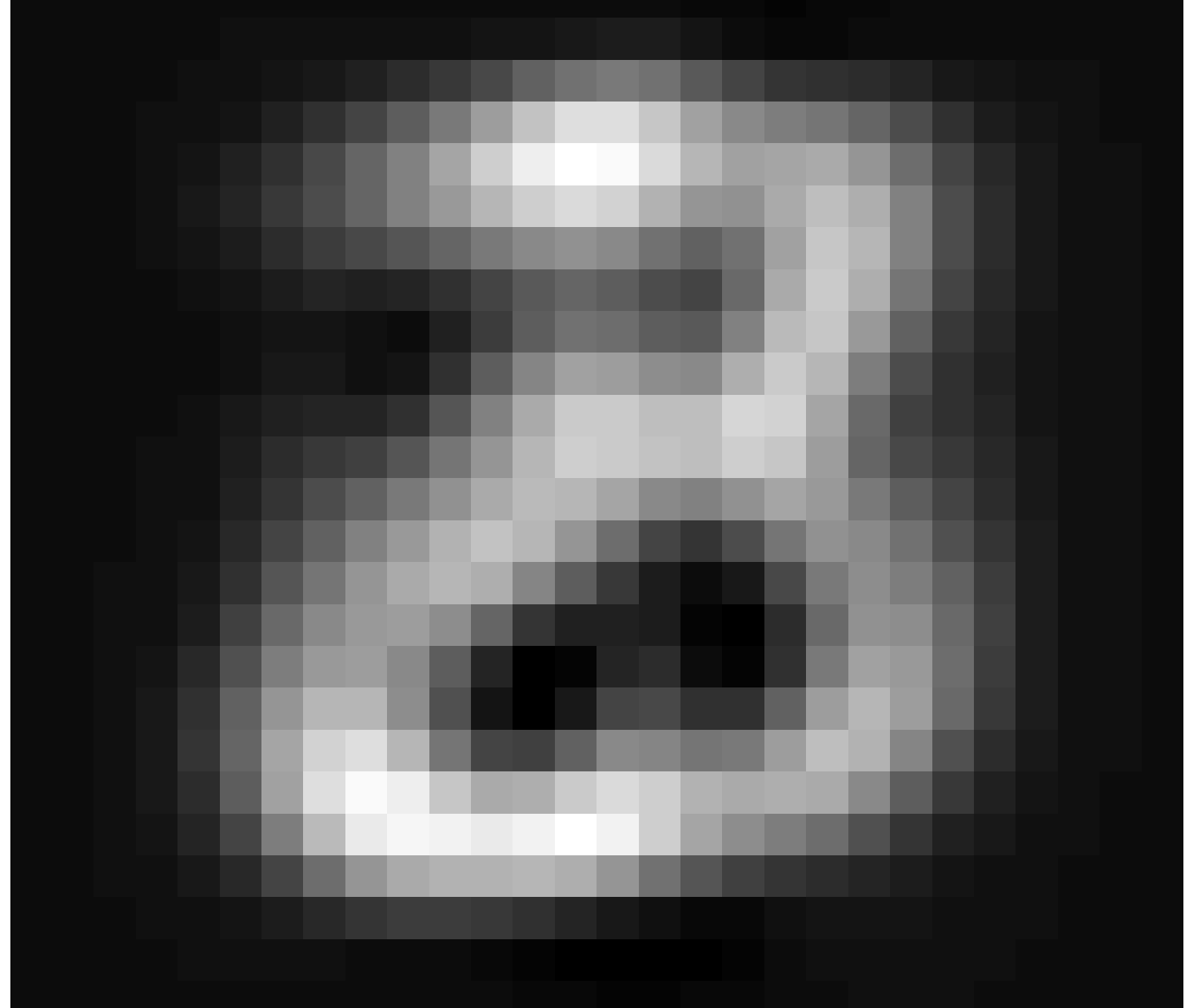}%
	\includegraphics[keepaspectratio, width=0.19\textwidth]{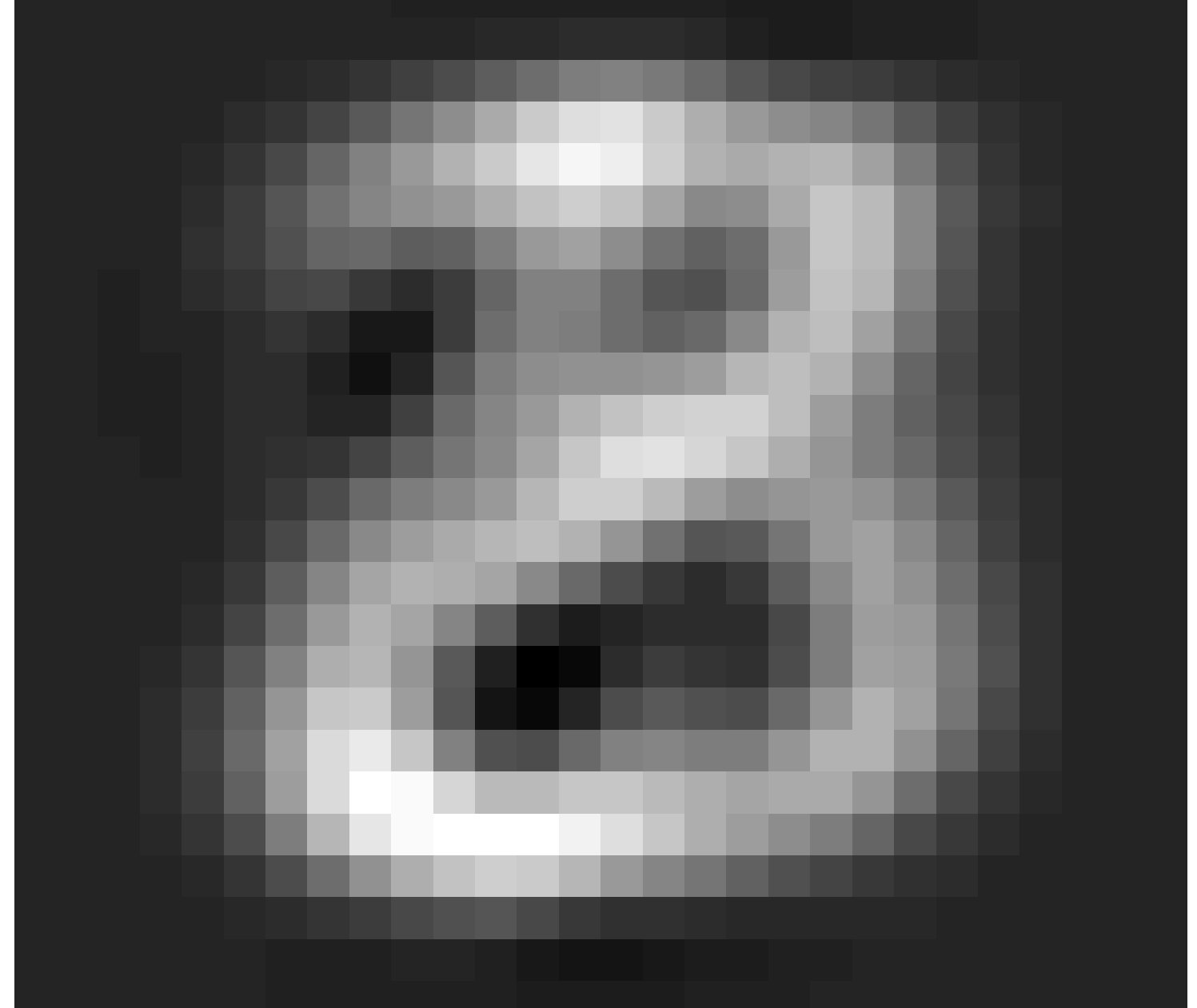}
	\caption{Experiments on MNIST dataset, whose original images are 
		reported in the first column. For each of those, we compute dropout for MF 
		with $\theta = 0.5$ and $\theta = 0.8$ - second and fourth columns 
		respectively - and the two relative closed form solutions \eqref{eq:Aopt} 
		- third and fifth columns. Additional digits in the Supplementary Material.} 
	\label{fig:zibibbo}
\end{figure*}

\begin{figure}[t!]
	\includegraphics[width=\columnwidth,keepaspectratio]{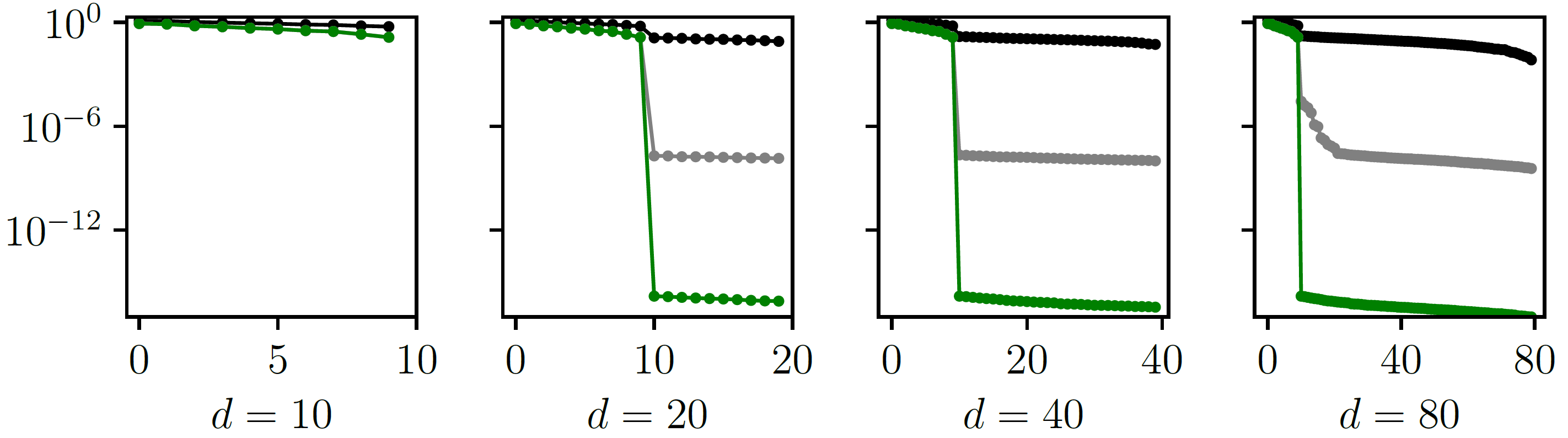}
	\caption{Singular values corresponding to the optimal solutions of the 
		three regularization schemes considered: fixed dropout rate of $\theta = 0.9$ 
		(black), 
		adaptive dropout $\theta = \theta(d)$ as \eqref{eq:theta_d} with $p = 0.9$ 
		(gray), and the nuclear-norm squared closed-form optimization as in 
		Proposition 
		\ref{prop:LowConvEnv} (green). 
		Best viewed in color.}
	\label{fig:rank_plots}
\end{figure}

\section{NUMERICAL SIMULATIONS}\label{sez:sim}

{\bf Stochastic vs. deterministic reformulations of 
dropout.} To demonstrate our claims experimentally, we first verify the 
equivalence between the stochastic \eqref{eq:dropoutMF} and its deterministic 
counterpart 
\eqref{eq:MF}, in which $\Omega = \Omd$. To do so, we construct a synthetic 
data matrix $\mathbf{X}$, where $m = n = 100$, 
defined as the matrix product $\mathbf{X} = \mathbf{U}_0 
{\mathbf{V}_0}^\top$ where $\mathbf{U}_0, \mathbf{V}_0 \in 
\mathbb{R}^{100 \times d}$ with $d=160$ (see the Supplementary Material for 
the cases $d=10,40$). The entries of 
$\mathbf{U}_0$ and $\mathbf{V}_0$ were sampled from a 
$\mathcal{N}(0,\varsigma^2)$ Gaussian 
distribution with standard deviation 0.1. Both the stochastic and 
deterministic formulations of dropout were solved by 10,000 
iterations of gradient descent with diminishing $O(\frac{1}{t})$ lengths for the 
step size. In the stochastic setting, we approximate the objective in 
\eqref{eq:dropoutMF} 
and the gradient by sampling a new Bernoulli vector $\mathbf{r}$ for every 
iteration of Algorithm \ref{alg:drop_tr}.

Figure \ref{fig:obj_curves} plots the objective curves for the stochastic and 
deterministic dropout formulations for different choices of the dropout rate $\theta 
= 0.1, 0.3, 0.5, 0.7, 0.9$ and factorization size $d = 10, 40, 160$. We observe 
that across all choices of parameters $\theta$ and $d$, the deterministic objective 
 \eqref{eq:MF} tracks the apparent expected value that is computed in
 \eqref{eq:dropoutMF}. This provides experimental evidence for the fact that the 
 two formulations are equivalent, as predicted.

{\bf Evaluating the connections with nuclear norm.} As a second experiment, we 
want to support the connection between $\Omd$ and the squared nuclear norm, in the case of a factorization with a variable size.


We constructed a synthetic dataset $X$ consisting of a low-rank matrix combined 
with dense Gaussian noise. Specifically, we let $X = U_0 V_0^\top + Z_0$ where 
$U_0, V_0 \in \mathbb{R}^{100 \times 10}$ contain entries drawn from a normal 
distribution $\mathcal{N}(0,\varsigma^2)$, with $\varsigma = 
0.1$. The entries of the noise matrix $Z_0$ were drawn from a normal distribution 
with $\varsigma = 0.01$. We fixed the dropout parameter $\bar{\theta} = 0.9$ 
and run Algorithm \ref{alg:drop_tr}.

Figure \ref{fig:rank_plots} plots the singular values for the optimal solution to each 
of the three problems. We observe first that without adjusting $\theta$, dropout 
regularization has little effect on the rank of the solution.  The smallest singular 
values are still relatively high and not modified significantly compared to the 
singular values of the original data. On the other hand, by adjusting the dropout 
rate based on the size of the factorization we observe that the method correctly 
recovers the rank of the noise-free data which also closely matches the predicted 
convex envelope with the nuclear-norm squared regularizer (note the log scale of 
the singular values). Furthermore, across the choices for $d$, the relative 
Frobenius distances between the solutions of these two methods are very small 
(between $10^{-6}$ and $10^{-2}$). Taken together, our theoretical predictions 
and experimental results suggest that adapting the dropout rate based on the size 
of the factorization is critical to ensuring the effectiveness of dropout as a 
regularizer and in limiting the degrees of freedom of the model.

{\bf Matrix factorization meets approximation with dropout.} In this paper, we 
study the process of dropping out columns of the 
factors $\U$ and $\V$ with which a data matrix $\X$ needs to be approximated in 
the form $\U\V^\top$. In addition to prove that this acts as a classical 
regularization scheme of the type \eqref{eq:MF}, we also show that, at the 
optimum, the same problem is equivalent with the matrix approximation 
framework \eqref{eq:MA}. As another experiment, we want to validate the quality 
of that approximation. In order to do this we consider MNIST training set, made of 
55K images of resolution 28$\times$28 that are vectorized and min-max 
normalized so that $\X$ has 55K rows and 784 columns. 

As a first step we fix $\theta$. Then, we applied SGD gradient descent, to compute 
the gradients 
as in Algorithm \eqref{alg:drop_tr} with a learning rate of $\epsilon = 10^{-4}$. In 
order to better cope with the non-convexity of the optimization, we performed 
about 1000 epochs where we carried 50$\times$ updates of $\U$ keeping $\V$ 
fixed and, conversely, 50$\times$ updates of $\V$ while freezing $\V$. Due to the 
shallowness of the model, we did not apply any batch strategy, but gradients are 
computed on the whole MNIST training by using acceleration with a GTX 
1080 GPU. We fixed the dimensionality of the factors to $40$.

While the factors $\U$ and $\V$ are computed in the aforementioned way, we 
compute the matrix $\U\V^\top$, dividing by $\theta$ and we compared against 
the closed form solution \eqref{eq:Aopt} of \eqref{eq:CLB}. In order to do so, we 
first compute 
$\gamma = \tfrac{1 - p}{p}$ being $p$ obtained by solving \eqref{eq:theta_d} 
with respect to $p$ while $\theta(d)$ and $d$ are fixed. Afterwards, we compute 
$\bar{d}$ as in \eqref{eq:bard} and, finally, we compute the singular value 
decomposition of $\X$ and we invoke \eqref{eq:Aopt} (in order to avoid 
out-of-memory issue, the svd of $\X$ was computed on a computer with 256 GB of 
RAM using MATLAB). In Figure \ref{fig:zibibbo} we show the visual results obtained 
comparing the original MNIST data with their reconstruction obtained through 
either dropout on MF or its convex lower bound. In both cases, we used two 
different dropout rates $\theta = 0.5$ and $\theta = 0.8$. Visually, the two 
reconstructions are pretty close and this is certified analytically since the mean 
reconstruction error of either dropout on MF or its convex lower bound has order of 
magnitude $10^{-2}$ and, the mean squared error between $\U\V^\top$ and 
\eqref{eq:Aopt} is approx. $10^{-3}$.


\section{CONCLUSIONS}\label{sez:end}

In this paper we present a theoretical analysis of dropout for matrix factorization 
(MF) \eqref{eq:dropoutMF}. In the case of a fixed size of the 
factors $d$, we proved that the expectation computed over $r_1,\dots,r_d \sim \Bt$ 
casts dropout for MF \eqref{eq:dropoutMF}  into the fully deterministic 
optimization problem \eqref{eq:MF} where $\Omega = \Omd$. For any fixed $d$, 
the two problems are equivalent in a very strong manner since, for any $\U$ and 
$\V$, the two objective functionals are point-wise equal and,
consequently, by either solving \eqref{eq:dropoutMF} or \eqref{eq:MF}, the 
optimal solution $\U^\opt$ and $\V^\opt$ is the same. 


Additionally we also showed a strong connection between nuclear norm regularization and dropout regularization.  In particular, we began by noting the close similarity between $\Omd$ and the variation form of the nuclear norm, but then we demonstrated that with a fixed choice of $\theta$ the resulting problem allows the size of factorization to grow unbounded.

We also investigated the case of a factorization with variable size. When $d$ varies, 
the regularizer $\Omd$ is pathologically promoting over-sized factorizations when 
$\theta$ is fixed. This motivated us in proposing an adapted choice for $\theta$ 
which, as defined in \eqref{eq:theta_d}, depends upon the size of the factorization 
$d$ and the hyper-parameter $p$. This stage ensures that, not only the 
aforementioned problem is solved, but at the same time, we are able to guarantee 
that $\theta = \theta(d)$ as in \eqref{eq:theta_d} prevents other issues to arise. 
This is true because we demonstrate that the lower convex bound of $\tfrac{1 - 
\theta(d)}{\theta(d)}\Omd$ is the nuclear norm squared. Ancillary, we took 
advantage of this result to prove that, the optimal dropout for MF factors 
immediately get for free the global optimum of the convex optimization problem 
\eqref{eq:CLB}. Since the latter is a convex (squared) nuclear norm regularization 
that, as we argumented, can be framed as an adaptive PCA that, also, learns from 
data the optimal size $\bar{d}$ \eqref{eq:bard} that should be used to reduce the 
dimensionality of the data. 

Additionally, our results show a novel interpretation of dropout that suggests it enforces spectral sparsity and thus acts to promote low-rank solutions.

Finally, we have verified our theoretical predictions via experiments on both simulated and real data, and our results suggest a novel approach to linear subspace learning which is worthy of further study in various applications for artificial intelligence.

\balance

\bibliographystyle{plain}
\bibliography{dropout}

\end{document}